\documentclass[lettersize,journal]{IEEEtran}
\usepackage{amsmath,amsfonts}
\usepackage{amssymb}
\usepackage{algorithmic}
\usepackage{algorithm}
\usepackage{array}
\usepackage[caption=false,font=normalsize,labelfont=sf,textfont=sf]{subfig}
\usepackage{textcomp}
\usepackage{stfloats}
\usepackage{url}
\usepackage{verbatim}
\usepackage{graphicx}
\usepackage{tikz}
\usepackage{makecell}
\usepackage{soul}
\usepackage{hyperref}
\usepackage{adjustbox}
\usepackage{booktabs}
\usepackage{xcolor}
\usepackage[numbers]{natbib}
\usepackage{multirow}
\newcommand{\ie}{i.e., }
\newcommand{\eg}{e.g., }

\definecolor{bestmodel}{HTML}{F2798D}
\newcommand{\first}[1]{\textcolor{bestmodel}{\textbf{#1}}}

\usepackage{pifont}
\newcommand{\cmark}{\ding{51}}%
\newcommand{\xmark}{\ding{55}}%

\begin{document}

\title{Deep learning for dynamic graphs: models and benchmarks}

\author{Alessio~Gravina$^*$ and Davide~Bacciu,~\IEEEmembership{Senior Member,~IEEE}
\thanks{$^*$ Corresponding Author}
\thanks{D. Bacciu and A. Gravina are with the Department
of Computer Science, University of Pisa, Italy. (e-mail: davide.bacciu@unipi.it and alessio.gravina@phd.unipi.it)\\Preprint. Under review}
}

\markboth{IEEE Transactions on Neural Networks and Learning Systems}%
{Gravina \MakeLowercase{\textit{et al.}}: Deep learning for dynamic graphs: models and benchmarks}


\maketitle

\begin{abstract}
Recent progress in research on Deep Graph Networks (DGNs) has led to a maturation of the domain of learning on graphs. Despite the growth of this research field, there are still important challenges that are yet unsolved. Specifically, there is an urge of making DGNs suitable for predictive tasks on 
real-world systems of interconnected entities
, which evolve over time. 

With the aim of fostering research in the domain of dynamic graphs, at first, we survey recent advantages in learning both temporal and spatial information, providing a comprehensive overview of the current state-of-the-art in the domain of representation learning for dynamic graphs. Secondly, we conduct a fair performance comparison among the most popular proposed approaches on node and edge-level tasks, leveraging rigorous model selection and assessment for all the methods, thus establishing a sound baseline for evaluating new architectures and approaches. 
\end{abstract}


\section{Introduction}
Graph representation learning has been gaining increasing attention over the recent years promoted by the ubiquitousness and expressiveness of structured relational information. Graphs are powerful tools to represent systems of relations and interactions, across several application fields where deep learning for graphs has found successful application, such as biology, social science and human mobility \cite{MPNN, bioinformatics, gravina_schizophrenia, gravina2023Covid, social_network, google_maps}.

The key challenge when learning from graph data is how to numerically represent the combinatorial structures for effective processing and prediction by the model. A classical predictive task of molecule solubility prediction, for instance, requires the model to encode both topological information and chemical properties of atoms and bonds. Graph representation learning 
solves the problem in a data-driven fashion, by \textit{learning} a mapping function that compresses the complex relational information of a graph into an information-rich feature vector that reflects both structural and label information in the original graph.

Despite the progress made in recent years in the field, which is mainly related to the family of Deep Graph Networks (DGNs)~\cite{BACCIU2020203, GNNsurvey}, a majority of the literature works focuses on networks that are \textit{static} snapshots of a phenomenon. Indeed, this is often a limitation when considering real-world processes, both natural and synthetic, where interactions evolve over time, \ie they are \textit{dynamic} in nature. Examples of such time-evolving systems can be found in social networks, where users can develop new friendships, citation networks, which are constantly updated with new publications, and in an e-commerce, where user behaviors change over time as well as their interactions with items. 

Representing a time-varying process through a static graph can be a reasonable choice in those scenarios in which the temporal dynamic is extremely slow, such as in a protein-protein interaction network. In general, however, ignoring the temporal information can harm the final performance of the predictor~\cite{T-GCN, tgn_rossi2020, dyrep, TGAT}. Therefore, the community has begun to look into \textit{dynamic graphs} and into models that can process the temporal dimension of a graph, as well as its spatial aspects. As a result, the last few years have witnessed a surge of works on dynamic graphs, leading to a fragmented and scattered literature with respect to model formalization, empirical setups and performance benchmarks. This aspect very much motivated us to look into a systematization of the literature which does not only look at surveying the existing works, but also actively promotes the identification of shared benchmarks and empirical protocols for the fair evaluation of  dynamic graph models. 

The present survey has a three-fold contribution. First, we propose a coherent formalization of the domain of representation learning for dynamic graphs, unifying different definitions and formalism gathered from the literature. Secondly, we provide a survey on representation learning for dynamic graphs under our unified formalism (the taxonomy behind our surveying methodology is depicted in Figure~\ref{fig:taxonomy}). Finally, we provide the graph learning community with a fair performance comparison among the most popular DGNs for dynamic graphs, using a standardized and reproducible experimental environment\footnote{We release openly the code at \url{https://github.com/gravins/dynamic_graph_benchmark}.}. Specifically, we performed experiments with a rigorous model selection and assessment framework, in which all models were compared using the same features, same datasets and the same data splits. As a by-product of our work, we also provide the community with a selection of datasets which we put forward as good candidates for the benchmarking of future works produced by the community.

Existing surveys on the topic of deep graph learning are \cite{dynamicgraph_survey, traffic_forecasting_survey}. Our novel contributions with respect to such works are: (i) a broader and more up-to-date coverage of literature; (ii) a benchmark and an empirical comparison between a broad range of methods; and (iii) the identification of a curated list of datasets in both discrete and continuous time setting. 

The remainder of the paper is organized as follows: Section~\ref{sec:static_graph} briefly surveys representation learning for \textit{static} graphs, providing general definitions and methods useful to define dynamic graph problems\footnote{The interested reader is referred to \cite{BACCIU2020203, GNNsurvey} for an in-depth analysis of the methods developed within the scope of representation learning for static graphs.}. Section~\ref{sec:dynamic_graph} formalizes representation learning for \textit{dynamic} graphs
, while Section~\ref{sec:D_TDG} and \ref{sec:C_TDG} survey the related literature. Section \ref{sec:benchmark} describes our empirical evaluation setting and provides an experimental comparison of most popular DGNs for dynamic graphs. Section~\ref{sec:conclusions} concludes the paper.

\tikzset{SG/.style={sibling distance = 2.5cm}}
\tikzset{DT/.style={sibling distance = 9cm}}
\tikzset{CT/.style={sibling distance = 3cm}}

\begin{figure*}[!]
\centering
\scalebox{0.61}{
\begin{tikzpicture}
[
    level 1/.style = {very thick, sibling distance = 12cm},
    edge from parent path=
{(\tikzparentnode.south) .. controls +(0,-0.3) and +(0,0.5)
                           .. (\tikzchildnode.north)}]
]

\node {\textbf{\textit{Deep Graph Networks}}}
    child {node {\textbf{\textit{Static graphs}}$^*$}
        child[SG] {node[align=center] {\textbf{\textit{Spectral}}\\\small{(Sec II.B.1)}\\}}
        child[SG] {node[align=center] {\textbf{\textit{Spatial}}\\\small{(Sec II.B.2)}\\\small{$^*$For a deeper analysis we refer to [7, 8]}}}
        child[SG] {node[align=center] {\textbf{\textit{Random walks}}\\\small{(Sec II.B.3)}\\}}
        edge from parent []
    } 
    child {node {\textbf{\textit{Dynamic graphs}}}
         child {node {\textbf{\textit{D-TDGs}}}
            child[DT]{node {\textbf{\textit{Spatio-temporal}}}
                 child[SG] {node[align=center] {\textbf{\textit{Stacked}}\\\small{(Sec IV.A.1)}}}
                 child[SG] {node[align=center] {\textbf{\textit{Integrated}}\\\small{(Sec IV.A.2)}}}
            }
            child[DT]{node {\textbf{\textit{Generic D-TDGs}}}
                 child [SG] {node[align=center] {\textbf{\textit{Integrated}}\\\small{(Sec IV.B.1)}}}
                 child [SG] {node[align=center] {\textbf{\textit{Stacked}}\\\small{(Sec IV.B.2)}}}
                 child [SG] {node[align=center] {\textbf{\textit{Meta}}\\\small{(Sec IV.B.3)}}}
                 child[SG] {node[align=center] {\textbf{\textit{Autoencoder}}\\\small{(Sec IV.B.4)}}}
                 child[SG] {node[align=center] {\textbf{\textit{Random walks}}\\\small{(Sec IV.B.5)}}}
            }
        }  
        child {node {\textbf{\textit{C-TDGs}}}
             child[SG] {node[align=center] {\textbf{\textit{Integrated}}\\\small{(Sec V.1)}}}
             child[SG] {node[align=center] {\textbf{\textit{Stacked}}\\\small{(Sec V.2)}}}
             child[SG] {node[align=center] {\textbf{\textit{Random walks}}\\\small{(Sec V.3)}}}
             child[SG] {node[align=center] {\textbf{\textit{Hybrid}}\\\small{(Sec V.4)}}}
            }
        };
\end{tikzpicture}
}
\caption{Taxonomy employed to structure our survey of TDG models.\label{fig:taxonomy}}
\end{figure*}
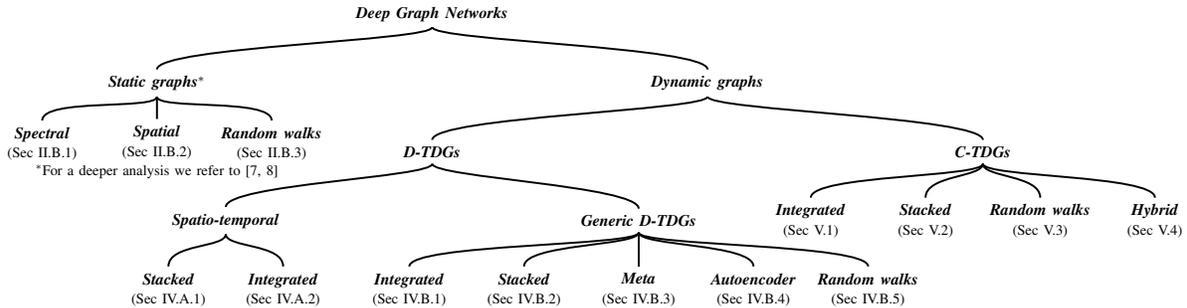

\section{Representation learning for \underline{static} graphs}\label{sec:static_graph}
\subsection{Definitions and notations}\label{sec:static_graph_notation}
A \textit{(static) graph} is a tuple $\mathcal{G}=(\mathcal{V}, \mathcal{E}, \mathbf{X}, \mathbf{E})$ defined by the nonempty set $\mathcal{V}$ of \textit{nodes} (also referred to as \textit{vertices}), and by the set $\mathcal{E}$ of \textit{edges} (also called \textit{links} or \textit{arcs})~\cite{graph_theory}. Nodes represent interacting entities, whereas edges denote connections between pairs of nodes. Depending on edge type, a graph is \textit{undirected}, when node pairs are unordered, \ie $\mathcal{E} \subseteq \{\{u,v\} \, | \, u,v \in \mathcal{V}\}$, or \textit{directed}, when the pairs are ordered, \ie $\mathcal{E} \subseteq \{(u,v) \, | \, u,v \in \mathcal{V}\}$. The structural information expressed by $\mathcal{E}$ can also be encoded into an \textit{adjacency matrix} $\mathbf{A}$, which is a square $|\mathcal{V}| \times |\mathcal{V}|$ matrix where each element $\mathbf{A}_{uv} \in \{0,1\}$ is $1$ if an edge connects the nodes $u$ and $v$, and it is $0$ otherwise.

\begin{figure}[ht]
\vspace{-0.55cm}
\centering 
\subfloat[]{\includegraphics[width=0.155\textwidth]{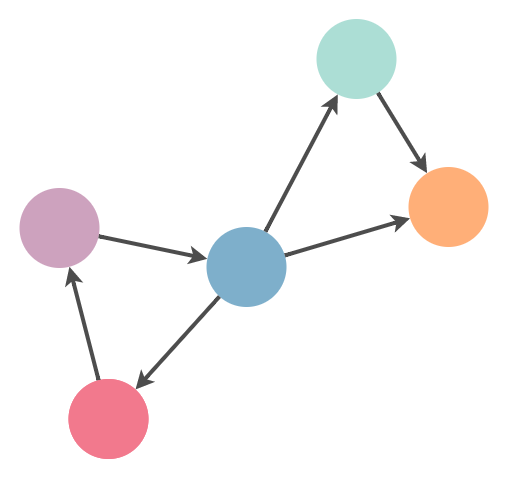}}
\hspace{0.1cm}
\subfloat[]{\includegraphics[width=0.155\textwidth]{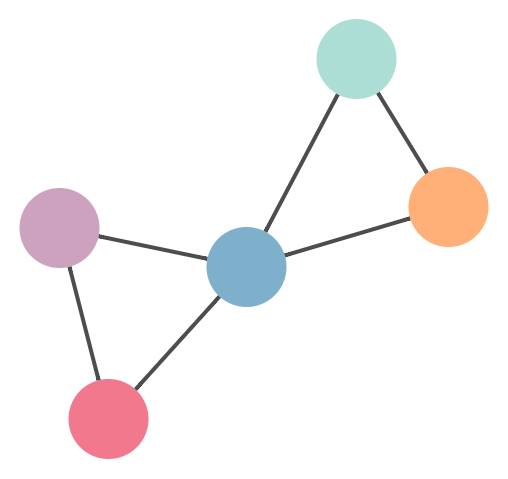}}
\subfloat[]{\includegraphics[width=0.159\textwidth]{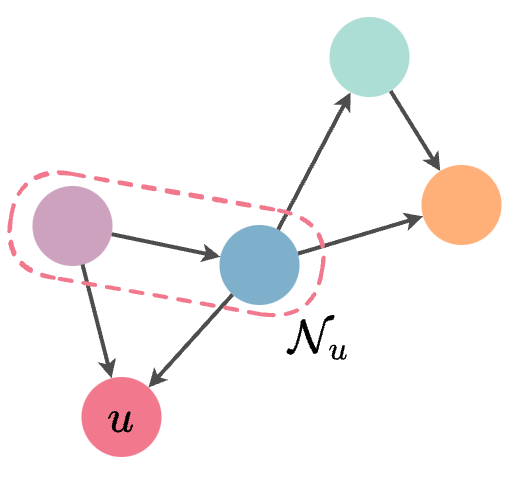}}
\caption{(a) A directed graph. (b) An undirected graph. (c) The neighborhood of node $u$.\label{fig:graph}}
\end{figure}

In many practical scenarios, nodes and edges are often enriched with additional attributes. We represent node features (also known as node representation or node embedding) as a matrix $\mathbf{X} \in \mathbb{R}^{|\mathcal{V}|\times d_n}$, where $d_n$ is the number of available features. The $u$-th row of $\mathbf{X}$ is denoted as $\mathbf{x}_v$ and represents a single node's features. Similarly, we represent edge features as a matrix $\mathbf{E} \in \mathbb{R}^{|\mathcal{E}| \times d_e}$, where $d_e$ is the number of edge features, and we indicate edge features vectors as $\mathbf{e}_{uv}$. Finally, we denote the neighborhood (or adjacency set) of a node $u \in \mathcal{V}$ as the set $\mathcal{N}_u = \{v\in\mathcal{V} \mid (v, u) \in \mathcal{E}\}$. A visual representation of a directed and undirected graph and the neighborhood of a node is shown in Figure~\ref{fig:graph}.

\subsection{Overview of representation learning for static graphs}
Representation learning for graphs has been pioneered by Graph Neural Network (GNN)~\cite{GNN} and Neural Network for Graphs (NN4G)~\cite{NN4G}, which were the first to provide learning models amenable also for cyclic and undirected graphs. The GNN leverages a \textit{recursive} approach, in which the state transition function updates the node representation through a diffusion mechanism that takes into consideration the current node and its neighborhood defined by the input graph. This procedure continues until it reaches a stable equilibrium. On the other hand, the NN4G leverages a \textit{feed-forward} approach where node representations are updated by composing representations from previous layers in the architecture. 

\begin{figure}[ht]
\vspace{-0.4cm}
\centering 
\includegraphics[trim={0 0 10mm 0}, clip, width=0.49\textwidth]{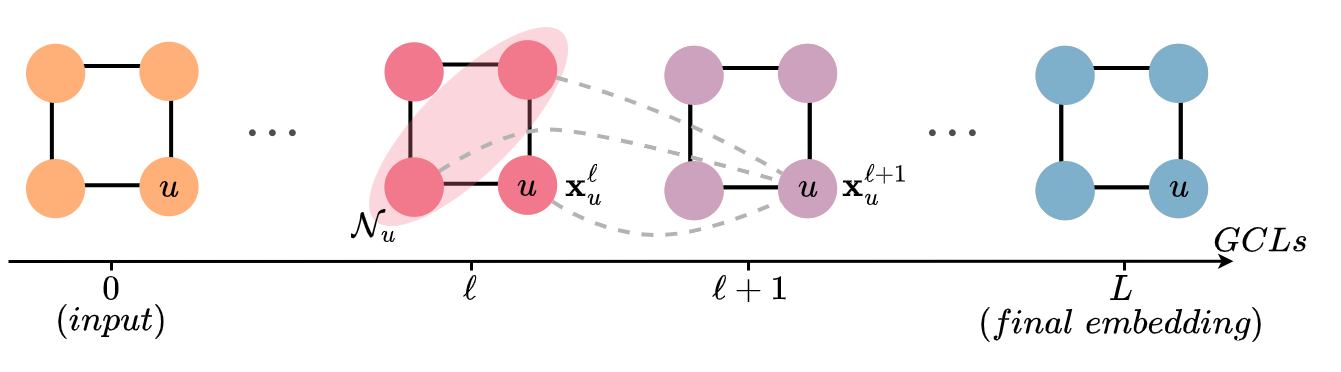}
\caption{Visual representation of a DGN. Given the input graph, each GCL $(\ell + 1)$ computes the new representation of a node $u$ as a transformation of $u$ and its neighbors representations at the previous layer, $\ell$.\label{fig:GCLs}}
\end{figure}

The original approaches by NN4G and GNN have been later extended and improved throughout a variety of approaches which can be cast under the umbrella term of (static) \textit{Deep Graph Networks} (DGNs), for which there exist dedicated surveys~\cite{BACCIU2020203, GNNsurvey}. Briefly, DGNs denote a family of approaches capable of learning the functional dependencies in a graph through a layered approach, where the single layers are often referred to as \textit{Graph Convolutional Layers} (GCLs). Each of these computes a transformation of node representations by combining the previous node representations and their neighborhoods. We visually represent this procedure in Figure~\ref{fig:GCLs}. The transformations are often referred to as \textit{graph convolutions}, and they are realized either in the \textit{spectral} or \textit{spatial} domain. 

\subsubsection{Spectral convolution}
In this setting, graphs are processed and learned through a parameterization in the spectral domain of their Laplacian matrices. Specifically, given a filter $\mathbf{g}_\theta = diag(\theta)$ parametrized by $\theta \in \mathbb{R}^{|\mathcal{V}|}$ and the graph signal $\mathbf{x} \in \mathbb{R}^{|\mathcal{V}|}$ for a graph $\mathcal{G}$, we can define the spectral graph convolution as a multiplication in the Fourier domain: 
\begin{equation}  
\label{eq:spectral_conv}
\small
    \mathbf{g}_\theta * \mathbf{x} = \mathbf{U} \mathbf{g}_\theta \mathbf{U}^T \mathbf{x}
\end{equation} 
where $\mathbf{U}^T \mathbf{x}$ is the graph Fourier transform, and $\mathbf{U}$ is the matrix of eigenvectors of the normalized graph Laplacian $\mathbf{L} = \mathbf{I} - \mathbf{D}^{-\frac{1}{2}}\mathbf{A}\mathbf{D}^{-\frac{1}{2}} = \mathbf{U}\Lambda \mathbf{U}^T$, with $\Lambda$ the diagonal matrix of the eigenvalue of $\mathbf{L}$. In the graph Laplacian, $\mathbf{I}$ indicates the identity matrix, $\mathbf{D}$ is the diagonal node degree matrix, and $\mathbf{A}$ is the adjacency matrix of $\mathcal{G}$. The approach in Equation~\ref{eq:spectral_conv} is severely limited by the computational requirements associated the Laplacian decomposition and by the spectral parameterization costs, which have motivated a whole body of followup works~\cite{chebnet, GCN}. Among these, the Graph Convolutional Network (GCN)~\cite{GCN} is certainly the most successful one.  
GCN leverages the degree-normalized Laplacian introduced in \cite{chebnet}, hence,
the output of the GCN's $(\ell+1)$-th layer for a node $u$ is computed as 
\begin{equation}
\small
    \label{eq:GCN}
    \mathbf{x}^{\ell+1}_u = \sigma\left(\mathbf{\Theta}_0 \mathbf{x}_u^{\ell} + \mathbf{\Theta}_1 \sum_{v \in \mathcal{N}_u} \frac{\mathbf{x}^{\ell}_v}{\sqrt{\mathbf{deg}(v)\,\mathbf{deg}(u)}}\right),
\end{equation}
where $\sigma$ is the activation function, while $\mathbf{deg}(v)$ and $\mathbf{deg}(u)$ are, respectively, the degrees of nodes $v$ and $u$. With such formulation, GCN requires $\mathcal{O}(|\mathcal{E}|)$ time.

\subsubsection{Spatial convolution}
Spatial convolutions are typically framed in the Message Passing Neural Network (MPNN)~\cite{MPNN} framework, where the representation for a node $u$ at a layer $\ell+1$ is computed as 
\begin{equation}
\small
    \label{eq:MPNN}
    \mathbf{x}_u^{\ell+1} = \phi_U(\mathbf{x}_u^\ell, \bigoplus_{v\in \mathcal{N}_u} \phi_M(\mathbf{x}_u^\ell, \mathbf{x}_v^\ell, \mathbf{e}_{uv}))
\end{equation}
where $\bigoplus$ is an aggregation invariant function, and $\phi_U$ and $\phi_M$ are respectively the \textit{update} and \textit{message} functions. The message function computes the message for each node, and then dispatches it among the neighbors. The update function collects incoming messages and updates the node state. A typical implementation of the MPNN use  sum  as $\bigoplus$ and $\phi_U$ functions, and $\phi_M(\mathbf{x}_u^\ell, \mathbf{x}_v^\ell, \mathbf{e}_{uv}) = \mathrm{MLP}(e_{uv})\mathbf{x}_v^\ell$. 

Depending on the definition of the update and message functions, it is possible to derive a variety of DGNs. The Graph Attention Network (GAT)~\cite{GAT} introduces an \textit{attention mechanism} to learn neighbors' influences
computing node representation as
\begin{equation}
\small
    \mathbf{x}_u^{\ell+1} = \sigma \left( \sum_{v \in \mathcal{N}_u} \alpha_{uv}\mathbf{\Theta} \mathbf{x}_v^\ell\right)    
\end{equation}
where $\alpha_{uv}$ is the classical softmax attention score between node $u$ and its neighbor $v$.

When graphs are large and dense, \ie $|\mathcal{E}|$close to $|\mathcal{V}|^2$, it can be impractical to perform the convolution over the node's neighborhood. Neighborhood sampling has been proposed as a possible strategy to overcome this limitation, i.e. by using only a random subset of neighbors to update node representation. GraphSAGE~\cite{SAGE} exploits this strategy to improve efficiency and scale to large graphs.
GraphSAGE updates the representation of a node $u$ by fixing the subset of nodes treated as neighbors, and by leveraging aggregation and concatenation operations:
\begin{equation}
\small
    \mathbf{x}_{u}^{\ell+1} = \sigma \Biggl(\mathbf{\Theta} \cdot \Bigl[\mathbf{x}_u^\ell \,\big\|\, \bigoplus_{v\in \mathcal{N}_{S}(u)}\mathbf{x}_v^\ell\Bigr]\Biggr)
\end{equation}
where $\mathcal{N}_S: \mathcal{V} \rightarrow \mathcal{V}$ is the function that computes the fixed subset of neighbors for a node $u$
. Differently, ClusterGCN~\cite{clusterGCN} samples a block of nodes identified by a graph clustering algorithm to restrict the neighborhood dimension.

The way models aggregate neighbors representations to compute node embeddings affects the discriminative power of DGNs. \citet{GIN} showed that most DGNs are at most as powerful as 1-Weisfeiler-Lehman test~\cite{WL}. In particular, Graph Isomorphism Network (GIN) \cite{GIN} has been proven to be as powerful as 1-Weisfeiler-Lehman test by computing node representations as 
\begin{equation}
\small
    \mathbf{x}_{u}^{\ell+1} = \mathrm{MLP} \left((1+\gamma)\mathbf{x}_u^\ell + \sum_{v \in \mathcal{N}_u} \mathbf{x}_v^\ell \right)
\end{equation}
with $\gamma$ as a learnable parameter or a fixed scalar.

More recently, advancements in the field of representation learning for graphs have introduced new architectures that establish a connection between the domains of DGNs and Ordinary Differential Equations (ODEs), with the primary objective of optimizing various aspects of message passing. These new methods exploit the intrinsic properties of ODEs to enhance the efficiency and effectiveness of message passing within DGNs. By formulating the propagation of information in graphs as an ODE system, these architectures effectively tackle multiple challenges, such as preserving long-range dependencies~\cite{gravina2023adgn}, reducing the computational complexity of message passing~\cite{dgc, sgc}, and mitigating the over-smoothing phenomena~\cite{pde-gcn, graphcon}.

\subsubsection{Random walks}
A different strategy to learn node embeddings including local and global properties of the graph relies on \textit{random walks}. A random walk is a random sequence of edges which joins a sequence of nodes. \citet{DeepWalk} proposed DeepWalk, a method that learns continuous node embedding by modeling random walks as the equivalent of sentences. 
Specifically, the approach samples multiple walks of a specified length for each node in the graph, and then it leverages the SkipGram model~\cite{skipgram} to update node representations based on the walks, treating the walks as sentences and the node representations as words within them.

Node2Vec~\cite{node2vec} improves DeepWalk by exploiting \textit{biased} random walks, \ie we can control the likelihood of revisiting a node in the walk (allowing the walk to be more or less explorative) and bias the exploration of new nodes towards a breath first or a depth first strategy.

\section{Representation learning for \underline{dynamic} Graphs: Notation and Taxonomy}\label{sec:dynamic_graph}
A \textit{dynamic graph} (also referred to as \textit{temporal} graph) is a tuple $\mathcal{G}(t)=(\mathcal{V}(t), \mathcal{E}(t), \mathbf{X}(t), \mathbf{E}(t))$, defined for $t\geq0$. Differently from static graphs, all  elements in the tuple are functions of time $t$. Thus, $\mathcal{V}(t)$ provides the set of nodes which are present in the graph at time $t$, and $\mathcal{E}(t) \subseteq \{\{u,v, t_i\} \, | \, u,v \in \mathcal{V}(t), t_i<t\}$
defines the links between them. Analogously, $\mathbf{X}(t)$ and $\mathbf{E}(t)$ define node states and edge attributes at time $t$. Although, $\mathcal{V}(t)$ can theoretically change over time, in practice it is often considered fixed for the ease of computation, which means that all the nodes that will appear in the dynamic graph are known in advance. Hence, $\mathcal{V}(t) = \mathcal{V}$ for $t\geq 0$.

The way we observe a system of interacting entities plays a crucial role in the definition of the corresponding dynamic graph. We can distinguish between two distinct types: \textit{discrete-time} dynamic graphs and \textit{continuous-time} dynamic graphs~\cite{dynamicgraph_survey}. Each of these representations gives rise to diverse architectures and learning approaches.
\\

A \textit{discrete-time dynamic graph} (D-TDG) models an evolving system that is fully observed at different timestamps. For such a reason, a D-TDG, $\mathcal{G} = \{\mathcal{G}_t \, | \, t\in[t_0, t_n]\}$, consists in a sequence of static graphs (known as \textit{snapshots}). Each snapshot, $\mathcal{G}_t = (\mathcal{V}_t, \mathcal{E}_t, \mathbf{X}_t, \mathbf{E}_t)$, provides a picture of the whole graph's state at a particular time $t$. Each snapshot maintains the notations and definitions outlined for static graphs (see Section~\ref{sec:static_graph_notation}).

We note that if the set of nodes and edges are fixed over time (\ie $\mathcal{G}_t = (\mathcal{V}, \mathcal{E}, \mathbf{X}_t, \mathbf{E}_t)$), then the dynamic graph is often referred to as \textit{spatio-temporal} graph.

\begin{figure*}[ht]
\centering 
\subfloat[]{\includegraphics[width=0.43\textwidth]{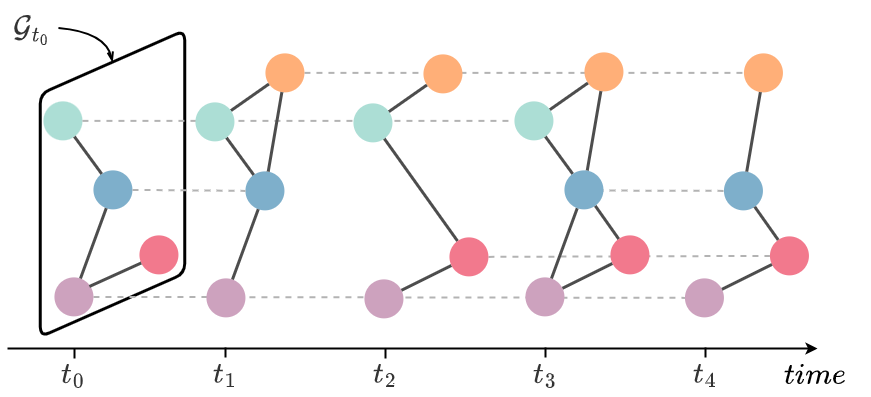}}
\subfloat[]{\includegraphics[width=0.43\textwidth]{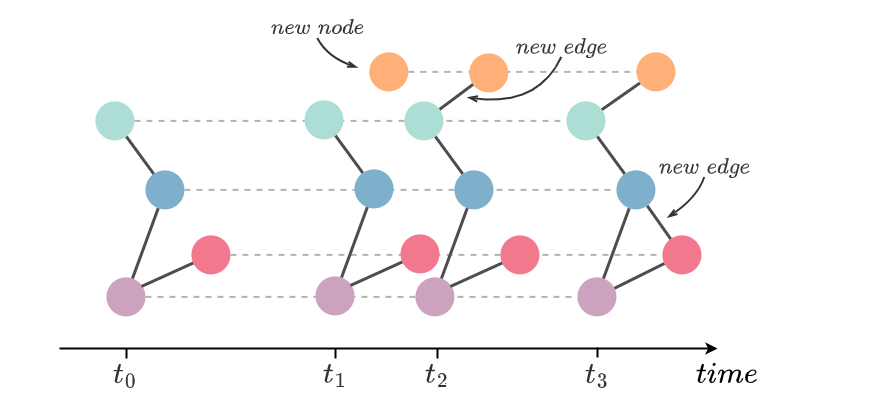}}
\caption{(a) A Discrete-Time Dynamic Graph defined over five timestamps and a set of five interacting entities. (b) The evolution of a Continuous-Time Dynamic Graph through the stream of events until the timestamp $t_3$.\label{fig:dyn_graph}}
\end{figure*}

Commonly, D-TDG are captured at periodic intervals (\eg hours, days,
etc.) Hence, considering $\Delta t > 0$ the interval between observations and $t_i$ the current timestamp, the next observation is captured at $t_{i+1} = t_i + \Delta t$. We present in Figure~\ref{fig:dyn_graph} a visual exemplification of a D-TDG.\\

A \textit{continuous-time dynamic graph} (C-TDG) is a more general formulation of a dynamic graph. 
It models systems that are not fully observed over time. In facts, only new events in the system are observed. Therefore, a C-TDG is a stream of events (observations) $\mathcal{G} = \{o_t\, | \, t\in[t_0, t_n]\}$. An event, $o_t = (t, \,EventType,\,\{u\}_{u \in \mathcal{V}(t)} )$, is a tuple containing the timestamp, the event type, and the nodes involved. 
Such events have a clear interpretation as graph edits~\cite{Gao2010}, which have also been formulated for dynamic graphs~\cite{paassen2021graph}. We consider six edit types: node deletions, node insertions, node replacements (i.e. features change), edge deletions, edge insertions, and edge replacements (i.e., edge features change). Without loss of generality, we can group these events into three main categories: \textit{node-wise} events, when a node is created or its features are updated; \textit{interaction} events, \ie a temporal edge is created; \textit{deletion} events, \ie node/edge is deleted.

Despite the underlying discrete nature of events,
the granularity of the observations is refined to the extent that the dynamic graph is considered as a continuous flow, allowing for events to happen at any moment (\ie characterized by irregular timestamps). This is in contrast to discrete-time dynamic graphs, where changes to the graph are typically aggregated.

Generally, the temporal neighborhood of a node $u$ at time $t$, consists of all the historical neighbors of $u$, prior to current time $t$, \ie $\mathcal{N}^t_u = \{(v, t_i) \, |\,  \{u,v,t_i\} \in \mathcal{E}(t), t_i < t  \}$. 

We observe that at any time point $t$, we can obtain a snapshot of the C-TDG by sequentially aggregating the events up to time $t$.  
Figure~\ref{fig:dyn_graph} shows visually the temporal evolution of a C-TDG.

We now proceed by providing a survey of state-of-the-art approaches in the domain of representation learning for dynamic graphs by means of the taxonomy and definition introduced in this section.

\section{Learning with Discrete-Time Dynamic Graphs}\label{sec:D_TDG}
Given the sequential structure of D-TDGs, a natural choice for many methods has been to extend Recurrent Neural Networks~\cite{RNN} to graph data. Indeed, most of the models presented in the literature can be summarized as a combination of static DGNs and RNNs. In particular, some approaches adopt a stacked architecture, where DGNs and RNNs are used sequentially, enabling to separately model spatial and temporal dynamics. Other approaches integrate the DGN inside the RNN, allowing to jointly capture the temporal evolution and the spatial dependencies in the graph. Thus, the primary distinction between static and dynamic approaches lies in their architectures. Static approaches predominantly utilize feedforward or recurrent architectures. Both used for exploring and learning the inherent static graph structure. Differently, dynamic approaches are characterized by recurrent architectures for learning and capturing temporal and spatial dependencies within the evolving graph, which mirrors the increased complexity inherent in addressing dynamic graphs. In the following, we review state-of-the-art approaches for both spatio-temporal graphs and more general D-TDGs.

\subsection{Spatio-temporal graphs}
When dealing with spatio-temporal graphs, new methods are designed to solve the problem of predicting the node states at the next step, $\mathbf{X}_{t+1}$, given the history of states, $\mathbf{X}_t$. 
To do so, different types of architectures have been proposed to effectively solve this task.

\subsubsection{Stacked architectures}
\citet{GCRN} proposed Graph Convolutional Recurrent Network (GCRN), one of the earliest deep learning models able to learn 
spatio-temporal graphs. 
The authors proposed 
to stack a Chebyshev spectral convolution~\cite{chebnet} (Equation~\ref{eq:GCN} shows the first-order approximation of this convolution) for graph embedding computation and a Peephole-LSTM~\cite{peephole-LSTM1, peephole-LSTM2} for sequence learning:
\begin{equation}\label{eq:GCRN}
\small
\begin{split}
    \mathbf{X}'_t &= \text{Cheb}(\mathbf{X}_t, \mathcal{E}, k, \mathbf{\Theta})\\
    \mathbf{H}_t &= \text{peephole-\textsc{lstm}}(\mathbf{X}'_t)
\end{split}
\end{equation}
where $\text{Cheb}(\mathbf{X}_t, \mathcal{E}, k, \mathbf{\Theta})$ represents Chebyshev spectral convolution (leveraging a polynomial of order $k$) computed on the snapshot $\mathcal{G}_t$ parametrized by $\mathbf{\Theta} \in \mathbb{R}^{k\times d_h\times d_n}$. Here, $d_h$ is the new latent dimension of node states, and $\mathbf{H}_t$ is the hidden state vector, which is equivalent to the node states at time $t+1$ (\ie $\mathbf{X}_{t+1}$). To ease readability, in the following, we
drop from the equation the edge set, $\mathcal{E}$, and the polynomial degree, $k$, since they are fixed for the whole snapshot sequence.

Equation~\ref{eq:GCRN} can be reformulated to define a more abstract definition of a stacked architecture between a DGN and an RNN, \ie
\begin{equation}\label{eq:stacked_architecture}
\small
\begin{split}
    \mathbf{X}'_t &= \text{\textsc{dgn}}(\mathbf{X}_t, \mathbf{\Theta})\\
    \mathbf{H}_t &= \text{\textsc{rnn}}(\mathbf{X}'_t)
\end{split}
\end{equation}

\citet{DCRNN} implement Equation~\ref{eq:stacked_architecture} by leveraging the same spectral convolution as GCRN and a Gated Recurrent Unit (GRU)~\cite{GRU} as RNN. Differently, \citet{T-GCN} employed the first-order approximation of the Chebyshev polynomials, which lead to the usage of a GCN to learn spatial features, and a GRU to extract temporal features. 
A3TGCN~\cite{a3tgcn} extends the implementation of \citet{T-GCN} with an attention mechanism to re-weight the influence of historical node states, with the aim of capturing more global information about the system.


\subsubsection{Integrated architectures}
In contrast to the aforementioned approaches, an alternative type of architecture is the one of an integrated architecture, where the DGN is incorporated into the RNN to simultaneously capture and integrate temporal evolution and spatial dependencies within the graph.
\citet{GCRN} proposed a second version of GCRN that exploit this type of architecture, by embedding the Chebyshev spectral convolution in the Peephole-LSTM. In this case, input, forget, and output gates can be reformulated as
\begin{equation}
\small
    \hat{h} = \sigma(\text{Cheb}(\mathbf{X}_t, \mathbf{\Theta}_{x}) + \text{Cheb}(\mathbf{H}_{t-1}, \mathbf{\Theta}_{h}) + \theta_{c} \odot c_{t-1}),
\end{equation}
the rest of the LSTM is defined as usual. We note that $\odot$ denotes the Hadamard product, $\sigma$ is the activation function, and $\hat{h}$ if the output of a generic gate. The weights $\mathbf{\Theta}_{h} \in \mathbb{R}^{k\times d_{h}\times d_{h}}$, $\mathbf{\Theta}_{x} \in \mathbb{R}^{k\times d_{h}\times d_{n}}$, $\theta_{c} \in \mathbb{R}^{d_{h}}$ 
are the parameters of the model. We observe that (here and in the following) the bias term is omitted for the easy of readability.

The Spatio-Temporal Graph Convolutional Network~\cite{STGCN} composes several spatio-temporal blocks to learn topological and dynamical features. Each block consists of two sequential convolution layers and one graph convolution in between. The temporal convolution layer contains a 1-D causal convolution followed by a Gated Linear Unit~\cite{GLU}, while the graph convolution is in the spectral domain. Let's consider $\mathrm{Conv}^\mathcal{G}$ the spectral graph convolution, and $\mathrm{Conv}^\mathcal{T}_1$ and $\mathrm{Conv}^\mathcal{T}_2$ the first and second temporal convolutions, respectively. Thus, each spatio-temporal block can be formulated as 
\begin{equation}
\small
    \mathbf{H}_t = \mathrm{Conv}^\mathcal{T}_2\left(\, \mathrm{ReLU}(\,\mathrm{Conv}^\mathcal{G}(\,\mathrm{Conv}^\mathcal{T}_1(\mathbf{X}_t)\,))\right).
\end{equation}

\citet{astgcn} extend the spatio-temporal blocks with an attention mechanism on both spatial and temporal dimensions to better capture the spatial-temporal dynamic of the graph.

\subsection{General D-TDGs}
Different from spatio-temporal graphs, the topology of a general D-TDG can evolve over time. In this case, the mere exploitation of the sole node states can lead to poor performance, since the new topology leads to different dynamics in the graph. In fact, the evolving topology is responsible for different information flows in the graph over time. Thus, excluding the evolution of the graph structure becomes a major limit of the method, leading to inaccurate predictions. 

Even in this case, we can categorize approaches for general D-TDGs depending on the architectural design.

\subsubsection{Integrated architectures}
\citet{GC-LSTM} proposed GC-LSTM an encoder-decoder model for link prediction, assuming fixed the node set $\mathcal{V}(t)$. The encoder consists of a GCN embedded in a standard LSTM. The GCN learns topological features of the cell state $c$ and of the hidden state $h$, which are used to save long-term relations and extract input information, respectively. The encoder takes as input the sequence of adjacency matrices and returns an embedding that encodes both temporal and spatial information. Thus, a generic gate in the LSTM can be expressed as:
\begin{equation}
\small
\hat{h}_t = \sigma(\mathbf{\Theta}_{h} \mathbf{A}_t + \text{\textsc{gcn}}_h(\mathbf{H}_{t-1}, \mathcal{E}_{t-1}))
\end{equation}
where $\mathbf{\Theta}_h \in \mathbb{R}^{|\mathcal{V}| \times d_{h}}$ is the weight matrix, and $\mathbf{A}_t$ is the adjacency matrix at time $t$, as usual. The decoder part of the model is a MLP that leverages the embedding generated from the encoder to predict the probability of each edge in future adjacency matrix $\mathbf{A}_{t+1}$. 

A similar strategy, which integrates topological changes into the computation, has also been employed by \citet{LRGCN}. 
Indeed, the authors proposed the so-called LRGCN that embed a Relational-GCN~\cite{RGCN} into a LSTM model. 
Different from GC-LSTM, LRGCN exploits the directionality of the edges in accordance with node features rather than the only stream of adjacency matrices, with the aim of effective modeling of the temporal dynamics. 
In LRGCN the input, forget, and output gates are computed as the result of the R-GCN model over the input node representations and the node embeddings computed at the previous step. The authors distinguish between 
four edge types to produce more informed latent representation: \textit{intra-incoming}, \textit{intra-outgoing}, \textit{inter-incoming}, \textit{inter-outgoing}. An inter-time relation corresponds to an arc $(u,v)$ present at the previous time $t-1$, while an intra-time relation is an arc $(u,v)$ present at current time $t$. To employ LRGCN in a \textit{path classification} task, the authors extend their model with a self-attentive path embedding (SAPE). 
Given the representations of $m$ nodes on a path, $P \in \mathbb{R}^{m \times d_o}$ with $d_o$ is the output dimension of LRGCN,
SAPE first applies LSTM to capture node dependency along the path sequence, \ie $\Gamma = \mathrm{LSTM}(P) \, \in \mathbb{R}^{m \times d_{new}}$. Then, SAPE uses the self-attentive mechanism to learn node importance and generate size-invariant representation of the path, \ie 
\begin{equation}
\small
    S=\mathrm{softmax}(\mathrm{MLP}(\mathrm{tanh}(\mathrm{MLP}(\Gamma))))  \, \in \mathbb{R}^{r \times m}
\end{equation} 
with $r$ an hyper-parameter. Lastly, the final path representation is obtained by multiplying $S$ with $\Gamma$, \ie $e=S\Gamma  \, \in \mathbb{R}^{r \times d_{new}}$.

With the aim of speeding up the dynamic graph processing, \citet{dyngesn} propose DynGESN, an extension of the Graph Echo State Network~\cite{gesn} to the temporal domain. Specifically, DynGESN updates the embedding for a node $u$ at time $t$ as
\begin{equation}
\small
    \mathbf{h}_{t}^u = (1-\gamma)\mathbf{h}_{t-1}^u + \gamma \text{tanh} \left( \mathbf{\Theta}_{i} \mathbf{h}^u_{t} +  \sum_{v \in \mathcal{N}^t_u}  \mathbf{\Theta}_{r} \mathbf{x}^v_{t-1} \right),
\end{equation}
with $0<\gamma\leq 1$  being a leakage constant, $\mathbf{\Theta}_{i}$ the input weights, and $\mathbf{\Theta}_{r}$ the recurrent weights. Both input and recurrent weights are randomly initialized.

\subsubsection{Stacked architectures}
Instead of integrating the DGN into the RNN, \citet{mpnn_lsltm} stack an LSTM on top of a MPNN, as previously proposed for spatio-temporal graphs. Differently from those approaches, the authors leverage as input the new node features as well as the new topology. Thus, the MPNN updates node representations by exploiting the temporal neighborhoods in each snapshot. 
\citet{roland} extend the MPNN-LSTM method by proposing the exploitation of \textit{hierarchical node states}. Thus, the authors propose to stack multiple DGN's layers and interleave them with the sequence encoder, \eg the RNN, to better exploit the temporal dynamic at each degree of computation. Thus, the node state at each layer depends on both the node state from the previous layer and the historical node state. More formally, the  $\ell$-th layer of \author{roland}'s framework is 
\begin{equation}
\small
\begin{split}
    \tilde{\mathbf{H}}^\ell_t &= \text{\textsc{dgn}}^\ell(\tilde{\mathbf{H}}_t^{\ell-1})\\
    \mathbf{H}_t^\ell &= \text{\textsc{update}}(\tilde{\mathbf{H}}^\ell_t, \mathbf{H}_t^{\ell-1}).
\end{split}
\end{equation}
where \textsc{update} is the sequence encoder. Similarly \citet{10.1145/3292500.3330919} employed GCN to learn spatial dependencies and MLP as update function.

Contrarily from previous works, \citet{cini2023scalable} propose to first embed the history of the node time series into latent representations that encode the temporal dynamic of the system. Such representations are then processed leveraging multiple powers of a graph shift operator (\eg graph laplacian or adjacency matrix) to encode the spatial dynamic of the system. Specifically, the authors propose to encode the temporal dynamic by means of an Echo State Networks (ESNs)~\cite{esn1, esn2}, a randomized recurrent neural networks\footnote{In randomized neural networks the hidden weights are randomly initialized and kept fixed after initialization. Only the weights in the final readout layer are learned, typically employing highly efficient methods like least-squared minimization~\cite{randomizedNN}.}, to efficiently compute node embedding and improve the scalability of DGNs for D-TDG.


\subsubsection{Meta architectures}
We refer to \textit{meta architectures} as those methods that learn a function that maps the evolution of the graph into the evolution of the parameters of the employed DGN. This kind of architecture has been proposed by \citet{egcn} to deal with those scenarios where nodes may frequently appear and disappear. As observed by the authors, %
such dynamics can be challenging to model with RNN-based models, since they have difficulties in learning these irregular behaviors. 
In such situation, the authors proposed Evolving GCN (E-GCN) to capture 
the dynamism of such graphs by using an RNN to evolve the parameters of a GCN. Thus, only the RNN parameters are trained. The authors considered two versions of their model, depending on whether graph structure or node features play the more important role. The first treats the GCN weights as the hidden state of a GRU to assign more significance to node representations. 
The second computes the weights as the output of the LSTM model, and it is more effective when the graph structure is important for the task. Let's consider $\mathrm{GRU}(\mathbf{X}_t, \mathbf{\Theta}_{t-1})$ as an extended version of a standard GRU model that exploits both the weight matrix at time $t-1$, $\mathbf{\Theta}_{t-1}$, and the previous node embedding, $\mathbf{H}_t$. The first E-GCN architecture can be formulated as
\begin{equation}
\small
\begin{split}
    \mathbf{\Theta}_{t} &= \mathrm{GRU}(\mathbf{X}_t, \mathbf{\Theta}_{t-1})\\
    \mathbf{H}_t &= \mathrm{GCN}(\mathbf{X}_t, \mathcal{E}_t, \mathbf{\Theta}_{t})
\end{split}
\end{equation}
while the second substitutes the GRU with an LSTM that takes as input only the weight matrix at time $t-1$.

\subsubsection{Autoencoder architectures}
\citet{dygrae} introduced DyGrAE, an autoencoder for D-TDGs. Specifically, DyGrAE leverages the Gated Graph Neural Network (GGNN)~\cite{gatedGNN} to capture spatial information, and LSTM encoder-decoder architecture to capture the dynamics of the network. GGNN is a DGN similar to the GNN introduced by \citeauthor{GNN}, but with a fixed number of iterations. DyGrAE consists of four components: a GGNN to learn the spatial dynamic; an RNN to propagate temporal information; an encoder to project the graph evolution into a fixed-size representation; and a decoder to reconstruct the structure of the dynamic graph. At each time step, 
at first, DynGrAE computes the snapshot embedding 
as the result of the average pooling on node embeddings at time $t$, \ie $emb(\mathcal{G}_t) = \mathrm{pool}_{avg}(\mathrm{GGNN}(\mathbf{X}_{t}))$. 
Then, the LSTM encoder-decoder uses the graph embeddings to encode and reconstruct the input graph sequence:
\begin{equation}
\small
    \begin{split}
        \mathrm{encoder}:\, \mathbf{h}^{enc}_t &= \mathrm{LSTM}_{enc} (emb(\mathcal{G}_t), \mathbf{h}^{enc}_{t-1})\\
        \mathrm{decoder}:\, \mathbf{h}^{dec}_t &= \mathrm{LSTM}_{dec} (\Bar{\mathbf{A}}_{t-1}, \mathbf{h}^{dec}_{t-1})
    \end{split}
\end{equation}
where $\Bar{\mathbf{A}}_{t-1} = \mathrm{sigmoid}(\mathrm{MLP}(\mathbf{h}^{dec}_{t-1}))$ is the reconstructed adjacency matrix at time $t-1$. The decoder uses $ h^{enc}_w$ to initialize its first hidden state, if $w$ is window size. To improve the performance, the authors introduced a temporal attention mechanism, which forces the model to focus on the time steps with significant impact. That mechanism causes the reformulation of the decoder as 
\begin{equation}
\small
    \mathbf{h}^{dec}_t = \mathrm{LSTM}_{dec} ([\mathbf{h}_{t}^* ||\Bar{\mathbf{A}}_{t-1}], \mathbf{h}^{dec}_{t-1})
\end{equation}
where $\mathbf{h}_{t}^* = \sum_{i=t-w}^{t-1} \Bar{\alpha}_t^i \mathbf{h}_i^{enc}$ is the attention distribution, the attention weights $\Bar{\alpha}_t^i = \mathrm{softmax}(f(\mathbf{h}^{dec}_{t-1}, \mathbf{h}^{enc}_{i}))$, and $f$ is a function, \eg dot product or MLP.
Dyngraph2vec~\cite{dyngraph2vec} introduces an analogous encoder-decoder approach featuring a deep architecture comprising dense and recurrent layers. This design facilitates the utilization of a more extended temporal evolution for predictions. 

A different strategy has been proposed by \citet{dyngem} that developed DynGEM. Such method handles D-TDGs by varying the size of the autoencoder network depending on a heuristic, which determines the number of hidden units required for each snapshot. Such heuristic, named PropSize, ensures that each pair of consecutive layers, $\ell$ and $\ell+1$, satisfy the condition:
\begin{equation}
\small
\label{eq:propsize}
    size(\ell+1) \geq \rho \cdot size(\ell)
\end{equation}
where $0<\rho<1$ is a hyper-parameter. This heuristic is applied to both encoder and decoder separately. If the condition in Equation~\ref{eq:propsize} is not satisfied for each pair of layers, then the number of $(\ell+1)$'s hidden units are increased. If PropSize is still unsatisfied between the penultimate and ultimate layers, a new layer is added in between. At each time step $t$ and before any application of PropSize, DynGEM initializes model parameters with those of the previous step $\mathbf{\Theta}_t = \mathbf{\Theta}_{t-1}$. This results in a direct transfer of knowledge between adjacent time steps, which guarantees a higher affinity between consecutive embeddings.

\subsubsection{Random walk based architectures}
Inspired by DeepWalk and Node2Vec, \citet{evolve2vec} propose a random walk approach for D-TDGs named Evolve2Vec. Given a sequence of graph snapshot, \citeauthor{evolve2vec} consider old interactions to contribute only in the propagation of topological information, while they use more recent interactions to encode the temporal dynamic. Thus, they proceed by aggregating old snapshots as a unique static graph.
Evolve2Vec starts random walks from all nodes with at least one outgoing edge in the static graph, as discussed in Section~\ref{sec:static_graph}. Then, in the temporal part, each walker move to a new neighbor if there is at least an outgoing edge in the current snapshot, otherwise it remains in the current node until an outgoing edge is added. Depending on how threshold between old and new is set, we can interpolate between a fully static or fully dynamic approach. After the computation of the random walks, node embeddings are computed by feeding the walks into a skip-gram model, as usual.

\section{Learning with Continuous-Time Dynamic Graphs}\label{sec:C_TDG}
In a scenario where the dynamic graph is observed only as new incoming events in the system, the methods defined in Section~\ref{sec:D_TDG} are unsuitable. In fact, approximating a C-TDG through a sequence of graph snapshots can introduce noise and loss of temporal information, since snapshots are captured at a more coarse level, with consequent performance deterioration. Moreover, the previously discussed methods usually do not allow including the time elapsed since the previous event. The majority of such methods update the embeddings only when new events occur. However, depending on how long it passed since the last event involving a node may result in the staleness of the embedding. 
%
Intuitively, the embedding may change depending on the time elapsed since the previous event. For such reasons, new techniques have been introduced to handle C-TDGs. We classify literature approaches into four categories depending on the architectural choices.

\subsubsection{Integrated architectures}
\citet{jodie} proposed JODIE, a method that learns embedding trajectories to overcome the staleness problem. JODIE computes the projection of a node $u$ in a future timestamp $t$ as an element-wise Hadamard product of the temporal attention vector with the previous node embedding,
\begin{equation}
\small
    \Hat{\mathbf{x}}_u(t) = (1+\mathbf{w}) \odot \mathbf{x}_u(t^-_u)
\end{equation}
where $(1+\mathbf{w})$ is the temporal attention vector, $\mathbf{w} = \mathbf{\Theta}_p\Delta t$ is the context vector, and $\Delta t = t - t^-_u$ is the time since the last event involving $u$. Thanks to the projection, JODIE can predict more accurately future embeddings, thus new events. Similar to other models, when an interaction event occurs between nodes $u$ and $v$, JODIE computes the embeddings $\mathbf{x}_u$ and $\mathbf{x}_v$ by leveraging two RNNs. 

\citet{dyrep} proposed DyRep, a framework that 
 update the representation of a node as it appear in an event in the C-TDG. DyRep captures the continuous-time dynamics leveraging a temporal point process approach. A temporal point process is characterized by the conditional intensity function that models the likelihood of an event to happen given the previous events. 
 DyRep's conditional intensity function, computed for an event between nodes $u$ and $v$ at time $t$, is:
\begin{equation}
\small
    \lambda_{uv}^k(t) = f_k(g^k_{uv}(t^-))
\end{equation}
where $k$ is the event type, $t^-$ is the previous timestamp in which an event occur, and 
\begin{equation}
\small
    f_k(z) = \psi_k \log\left(1 + \exp\left(\frac{z}{\psi_k}\right)\right)
\end{equation}
with $\psi_k$ a parameter to be learned. The inner function 
\begin{equation}
\small
    g^k_{uv}(t^-) = \omega_k^T \cdot [\mathbf{x}_u(t^-) || \mathbf{x}_v(t^-)] 
\end{equation}
is a function of node representations learned through a DGN, with $\omega_k \in \mathbb{R}^{2|F|}$ the model parameters that learn time-scale specific compatibility. Node embeddings computed by the DGN are updated as
\begin{equation}
\small
\label{eq:dyrep_dgn}
    \mathbf{h}_u(t) = \sigma(\mathbf{\Theta}_{i} \mathbf{h}_u^{loc}(t^-) + \mathbf{\Theta}_{r} \mathbf{h}_u(t^-_u) + \mathbf{\Theta}_{e}(t - t^-_u))
\end{equation}
where $h_u^{loc}(t^-) \in \mathbb{R}^{d_h}$ is the representation of the aggregation of $u$'s direct neighbors, $t^-_u$ 
is the timestamp of the previous event involving node $u$, and $\mathbf{\Theta}_{i}, \mathbf{\Theta}_{r}  \in \mathbb{R}^{d_h\times d_h}$ and $\mathbf{\Theta}_{e} \in \mathbb{R}^{d_h}$ are learnable parameters. In Equation~\ref{eq:dyrep_dgn} the first addend propagates neighborhood information, the second self-information, while the third considers the exogenous force that may smoothly update node features during the interval time. To learn $\mathbf{h}_u^{loc}(t^-)$, DyRep uses an attention mechanism similar to the one proposed in the GAT model by \citet{GAT}. In this case, the attention coefficient is parametrized by $\mathcal{S} \in \mathbb{R}^{|\mathcal{V}| \times |\mathcal{V}|}$, which is a stochastic matrix denoting the likelihood of communication between each pair of nodes. $\mathcal{S}$ is updated according to the conditional intensity function. The aggregated neighborhood representation is 
\begin{equation}
\small
    \mathbf{h}_u^{loc}(t^-) = \max(\{\sigma(\alpha_{uv}(t) \cdot \mathbf{h}_v(t^-)) \,|\, v \in \mathcal{N}_u^t) \}),
\end{equation}
with $\sigma$ the activation function and $\alpha_{uv}(t)$ the attention factor, as usual.

\subsubsection{Stacked architectures}
In the case of sequential encoding of spatial and temporal information, \citet{TGAT} introduce 
TGAT, a model that learns the parameters of a continuous function that characterize the continuous-time stream. Similar to GraphSAGE and GAT models, TGAT employs a local aggregator that takes as input the temporal neighborhood and the timestamp and computes a time-aware embedding of the target node by exploiting an attention mechanism. 
The $\ell$-th layer of TGAT computes the temporal embedding of node $u$ at time $t$ as
\begin{equation}
\small
    \mathbf{h}^\ell_u(t) = \text{\textsc{mlp}}^\ell_2(\mathrm{ReLU}(\text{\textsc{mlp}}^\ell_1([\hat{\mathbf{h}}(t) || \mathbf{x}_u])))
\end{equation}
where $\hat{\mathbf{h}}(t)$ is the attentive hidden neighborhood representation obtained as
\begin{equation}
\small
    \begin{split}
        \mathbf{q}(t) &= \left[\mathbf{Z}(t)\right]_0 \mathbf{\Theta}_q\\
        \mathbf{K}(t) &= \left[\mathbf{Z}(t)\right]_{1:n} \mathbf{\Theta}_K\\
        \mathbf{V}(t) &= \left[\mathbf{Z}(t)\right]_{1:n} \mathbf{\Theta}_V\\
        \hat{\mathbf{h}}(t) &= \mathrm{attn}(\mathbf{q}(t), \mathbf{K}(t), \mathbf{V}(t))
    \end{split}
\end{equation}
where $\mathbf{Z}(t)= \left[\mathbf{x}^{\ell-1}_u(t) || \Phi_d(0), ..., \mathbf{x}^{\ell-1}_{v}(t) || \Phi_d(t-t_v)\right] \in \mathbb{R}^{(d_{h}+d_{t}) \times n}$ is the temporal feature matrix that concatenates the representation of each node in the neighborhood of $u$ with the time difference between the current time and the time of the previous event involving the neighbor, with $v\in \mathcal{N}_u^t$ and $n$ the size of $u$'s neighborhood; 
 $\mathbf{q}(t)$, $\mathbf{K}(t)$, and $\mathbf{V}(t)$ are the query, key and value projections of the matrix; and $\mathrm{attn}$ is an attention mechanism similar to GAT. The dimensional functional mapping $\Phi_d: t \rightarrow \mathbb{R}^{d_t}$ is defined as
\begin{equation}\label{eq:time_enc}
\small
    \Phi_d(t) = [\cos(\omega_1t)\sin(\omega_1t), ..., \cos(\omega_dt)\sin(\omega_dt)]
\end{equation}
where $\omega_i$ are learnable parameters.

Differently, \citet{streamgnn} proposed an approach, named StreamGNN, to learn the node embedding evolution as new edges appear in the dynamic graph. Thus, it is design to only deal with interaction events.  StreamGNN is composed of two main components: the update component, which is responsible for updating the node representations of the source and destination nodes of the new link; and the propagation component, which propagates the new event across the direct neighborhood of the involved nodes. When a new event is observed, the update component computes the representation of the event as the result of an MLP on the node representation of both source and destination. Then, such representation is updated by an LSTM to include historical information from previous interactions. The amount of the past node history used by the LSTM is inversely proportional to the time difference with the previous node interaction. Then, the lastly computed node embeddings of source and target nodes are merged with the output of the LSTM model.  
After these first steps, the propagation component diffuse the computed representations across the 1-hop neighborhood by leveraging an attention mechanism and by filtering out those neighbors which appear in an interaction before a predefined threshold.

\citet{tgn_rossi2020} extend previous concepts by proposing a general framework composed of five core modules: memory, message function, message aggregator, memory updater, and the embedding module. The memory at time $t$ is a matrix $\mathbf{s}(t)$ that has the objective of representing the node's history in a vectorial format. For this purpose, it is updated after every event. The message function has the role of encoding the event to update the memory module. Given an interaction event involving nodes $u$ and $v$ at time $t$, the message function computes two messages
\begin{equation}
\small
    \begin{split}
    m_u(t) &= \mathrm{msg}_{src}(\mathbf{s}_u(t^-), \mathbf{s}_v(t^-), \Delta t,  \mathbf{e}_{uv}(t))\\
    m_v(t) &= \mathrm{msg}_{dst}(\mathbf{s}_v(t^-), \mathbf{s}_u(t^-), \Delta t,  \mathbf{e}_{uv}(t)),
    \end{split}
\end{equation}
where $\mathrm{msg}$ can be any learnable function, \eg a MLP. In case of a node event, it is sent a single message. The message aggregator is a mechanism to aggregate messages computed at different timestamps. It can be a learnable function, \eg RNN, or not, \eg message average or most recent message. After every event involving a node $u$, the memory of the node is updated by the memory updater as
\begin{equation}
\small
    \mathbf{s}_u(t) = \mathrm{mem}(\Bar{m}_u(t), \mathbf{s}_u(t^-))
\end{equation}
where $\Bar{m}_u(t)$ represents the 
aggregation of computed messages in a batch related to node $u$, and $\mathrm{mem}$ is an RNN. Lastly, the embedding module generates the representation for a node $u$ at time $t$ by exploiting the information stored in the memory module of the node itself and its neighborhood up to time $t$
\begin{equation}
\small
    \mathbf{h}_u(t) = \sum_{v \in \mathcal{N}_u^t} f(\mathbf{s}_u(t),  \mathbf{s}_v(t),  \mathbf{x}_u(t),  \mathbf{x}_v(t),  \mathbf{e}_{uv})
\end{equation}
with $f$ a learnable function and $\mathbf{x}_u(t)$, $\mathbf{x}_v(t)$ the input node representations of nodes $u$ and $v$.\\

\subsubsection{Random walk based architectures}
Even in the scenario of C-TDGs, it is possible to compute node embeddings relying on random walks. Differently from a standard random walk, in the continuous-time domain 
a valid walk is a sequence of interaction events with a non-decreasing timestamp. \citet{temporal_node2vec} extended the Node2Vec framework to exploit temporal random walks. Once decided the starting timestamp $t_0$, which is used to \textit{temporally bias} the walk, the framework samples new nodes for the walk by considering the temporal neighborhood. Differently from the general formulation of temporal neighborhood, \citeauthor{temporal_node2vec} apply a threshold to discriminate and filter old neighbors. 
The distribution to sample nodes in the walk can be either uniform, \ie $\mathbb{P}(v) = 1/|\mathcal{N}^t_u|$, or biased. Specifically, the authors proposed two ways to obtain a temporally weighted distribution. Let consider that the random walk is currently at the node $u$. In the first case, a node $v$ is sampled with the probability
\begin{equation}
\small
    \mathbb{P}(v) = \frac{\exp(\mathcal{T}(v) - \mathcal{T}(u))}{\sum_{v' \in \mathcal{N}^t_u}\exp(\mathcal{T}(v') - \mathcal{T}(u))},
\end{equation}
where $\mathcal{T}: \mathcal{V} \rightarrow \mathbb{R}^+$ is the function that given a node return the corresponding timestamp of the event in which the node was involved; while in the second
\begin{equation}
\small
    \mathbb{P}(v) = \frac{\delta(v, \mathcal{T}(v))}{\sum_{v' \in \mathcal{N}^t_u}\delta(v', \mathcal{T}(v'))},
\end{equation}
where $\delta : \mathcal{V}\times \mathbb{R}^+ \rightarrow \mathbb{Z}^+$ is a function that sorts temporal neighbors in descending order depending on time, thus returns a score that biases the distribution towards the selection of edges that are closer in time to the current node.

Instead of temporal random walks, \citet{CAW} exploited \textit{Causal Anonymous Walks} (CAW) to model C-TDGs. A CAW encodes the causality of network dynamics by starting from an edge of interest and backtracking adjacent edges over time. Moreover, a CAW is anonymous because it replaces node identities in a walk with relative identities based on the appearance order. The causality extraction helps the identification of temporal network motif, while node anonymization guarantees inductive learning. Given an edge $\{u,v\}$, the model extracts $M$ walks of length $m$ starting from both $u$ and $v$, and then performs the anonymization step. Afterward, an RNN encodes each walks leveraging two functions. The first consists of two MLPs ingested with the encoding of the correlation between the node $w$ and the sampled walks
\begin{equation}
\small
    f_1(w) = \mathrm{MLP}(g(w, S_u)) + \mathrm{MLP}(g(w, S_v))
\end{equation}
where $S_u$ is the set of sampled walks started from $u$, and $g$ is the function that counts the times a node $w$ appears at certain positions in $S_u$. The second function encodes time as Equation~\ref{eq:time_enc}. All the encoded walks are aggregated through mean-pooling or the combination of self-attention and mean-pooling to obtain the final edge representation.

NeurTWs~\cite{neurtws} extends \citet{CAW} by employing a different sampling strategy for the temporal random walks, which integrates temporal constraints, topological properties, and tree traversals, allowing to sample \textit{spatiotemporal-biased random walks}. These walks prioritize neighbors with higher connectivity (promoting exploration of more diverse and potentially expressive motifs), while being aware of the importance of recent neighbors. Furthermore, NeurTWs replaces the RNN-based encoding approach for walks with a component based on neural ODEs to facilitate the explicit embedding of irregularly-sampled events. \\

\subsubsection{Hybrid architectures}
\citet{pint} propose to improve the expressive power of methods designed for C-TDGs by leveraging the strengths of both CAW and TGN-based architectures. Thus, by providing a hybrid architecture.
Specifically, the authors observe that for TGN-based architectures, most expressive power is achieved by employing injective embedding module, message aggregator and memory updater functions. On the other hand, the main advantage of CAW is its ability to leverage node identities to compute representative embeddings and capture correlation between walks. However, such approach imposes that walks have timestamps in decreasing order, which can limit its ability to distinguish events. Under such circumstances, the authors propose PINT, an architecture that leverages injective temporal message passing and relative positional features to improve the expressive power of the method. Specifically, the embedding module computes the representation of node $u$ at time $t$ and layer $\ell$ as
\begin{equation}
\small
\begin{split}
    \hat{\mathbf{h}}^\ell_u(t) &= \sum_{v \in \mathcal{N}_u^t} \text{\textsc{mlp}}^\ell_{agg}(\mathbf{h}^{\ell-1}_v(t) || \mathbf{e}_{uv}) \alpha^{-\beta(t-t^-)}\\
    \mathbf{h}^\ell_u(t) &=\text{\textsc{mlp}}^\ell_{upd}(\mathbf{h}^{\ell-1}_u(t) || \hat{\mathbf{h}}^\ell_u(t))
\end{split}
\end{equation}
where $\alpha$ and $\beta$ are scalar hyper-parameters the node state is initialized with its memory representation, \ie $\mathbf{h}^{0}_u(t)=\mathbf{s}_u(t)$. To boost the power of PINT, the authors augment memory states with relative positional features, which include information about the number of existing temporal walks of a given length between two nodes.
\section{The benchmark problem}\label{sec:benchmark}

In this section, we provide the graph learning community with a performance comparison among the most popular DGNs for dynamic graphs. The aim is to support the tracking of the progress of the state-of-the-art and to provide robust baseline performance for future works. To the best of our knowledge, in fact, there are no widely agreed standard benchmarks in the domain of dynamic graphs. For such a reason, nowadays, it is not easy to fairly compare models presented in different works, because they typically use different data and empirical settings. The latter plays a crucial role in the definition of a fair and rigorous comparison, \eg including multiple random weights initialization and hyper-parameter search and similar data splits.

With this in mind, we designed three benchmarks to assess models that deal with spatio-temopral graphs, general D-TDGs, and C-TDGs.
Specifically, we evaluated methods for D-TDGs on both link and node prediction tasks while we constrained the evaluation of C-TDG methods to link prediction tasks due to the scarcity of suitable datasets.
To do so, we extended the library PyDGN~\cite{pydgn} to the D-TDG learning setting to foster reproducibility and robustness of results. With the same aim, we developed a Pytorch Geometric~\cite{Fey/Lenssen/2019} based framework to allow reproducible results in the continuous scenario. Lastly, 
in 
the Supplementary materials (Appendix~\ref{app:a}) we provide the community with a selection of datasets useful for benchmarking future works. An interest reader is referred to SNAP~\cite{snapnets}, TSL~\cite{tsl}, TGB~\cite{TGB}, and Network Repository~\cite{nr} for a broader data collections.

\subsection{Spatio-temopral graph benchmark}
In the spatio-temporal setting, we consider three graph datasets for traffic forecasting, \ie Metr-LA~\cite{DCRNN}, Montevideo~\cite{rozemberczki2021pytorch}, and PeMSBay~\cite{DCRNN}, and Traffic~\cite{LRGCN}. Specifically, 
\begin{itemize}
    \item \textbf{Metr-LA} consists of four months of traffic readings collected from 207 loop detectors in the highway of Los Angeles County every five minutes;
    \item \textbf{Montevideo} comprises one month of hourly passenger inflow at stop level for eleven bus lines from the city of Montevideo;
    \item \textbf{PeMSBay} contains six months of traffic readings collected by California Transportation Agencies (CalTrans) Performance Measurement System (PeMS) every five minutes by 325 traffic sensors in San Francisco Bay Area.
    \item \textbf{Traffic}: consists of traffic data collected over a period of three months, with hourly granularity, from 4,438 sensor stations located in the 7th District of California.
\end{itemize}
For all the three datasets, the objective is to perform \textit{temporal node regression}, thus, to predict the future node values, $\mathbf{X}_{t+1}$, given the past graph history, $[\mathcal{G}_i]_{i=1}^t$.

The baseline performance for this type of predictive problems on graphs is based on five spatio-temporal DGNs (\ie A3TGCN~\cite{a3tgcn}, DCRNN~\cite{DCRNN}, GCRN-GRU~\cite{GCRN}, GCRNN-LSTM~\cite{GCRN}, TGCN~\cite{T-GCN}), within the aim of assessing both stacked and integrated architectures, and the influence of an attention mechanism.

We designed each model as a combination of three main components. The first is the encoder which maps the node input features into a latent hidden space; the second is the DGN which computes the spatio-temporal convolution; and the third is a readout that maps the output of the convolution into the output space. The encoder and the readout are MLPs that share the same architecture among all models in the experiments.
We performed hyper-parameter tuning via grid search, optimizing the Mean Absolute Error (MAE). We perform a time-based split of the dataset which reserves the first 70\% of the data as training set, 15\% of the following data as validation set, and the last 15\% as test set. We trained the models using Adam optimizer for a maximum of 1000 epochs and early stopping with patience of 50 epochs on the validation error. For each model configuration, we performed 5 training runs with different weight initialization and report the average of the results. We report in Table~\ref{tab:st_grid} the grid of hyper-parameters exploited for this experiment. 
\renewcommand{\arraystretch}{1.4}
\begin{table}[ht]
\centering
\caption{The grid of hyper-parameters employed during model selection for the spatio-temporal tasks. \label{tab:st_grid}}
\scriptsize
\begin{tabular}{ll}
\specialrule{.15em}{.05em}{.05em} 
 & \textbf{Configs}\\\hline
learning rate & $10^{-2}$, $10^{-3}$, $10^{-4}$\\
weight decay & $10^{-3}$, $10^{-4}$\\
embedding dim & 1, 2, 4, 8\\
$\sigma$ & ReLU\\
Chebishev poly. filter size & 1, 2, 3\\
normalization scheme for $\mathbf{L}$ & \makecell[l]{$\mathbf{L} = \mathbf{D} - \mathbf{A}$,\\$\mathbf{L} = \mathbf{I} - \mathbf{D}^{-1/2} \mathbf{A} \mathbf{D}^{-1/2}$,\\$\mathbf{L} = \mathbf{I} - \mathbf{D}^{-1} \mathbf{A}$} \\
\specialrule{.15em}{.05em}{.05em} 
\end{tabular}
\end{table}

\paragraph*{Results}
In Table~\ref{tab:spatio-temp} we report the results on the spatio-temporal-based experiments. 
\renewcommand{\arraystretch}{1.4}
\begin{table*}[ht]
\centering
\caption{Mean test scores of the spatio-temporal models and std averaged over 5 random weight initializations. MAE is the optimized metric. The lower, the better.}\label{tab:spatio-temp}
\scriptsize
\begin{tabular}{lcc|cc|cc|cc}
\specialrule{.15em}{.05em}{.05em} 
 & \multicolumn{2}{c}{\textbf{Montevideo}} & \multicolumn{2}{c}{\textbf{Metr-LA}} & \multicolumn{2}{c}{\textbf{PeMSBay}}  & \multicolumn{2}{c}{\textbf{Traffic}}  \\
\textbf{Model} & \textbf{MAE} & \textbf{MSE} & \textbf{MAE} & \textbf{MSE} & \textbf{MAE} & \textbf{MSE} & \textbf{MAE} & \textbf{MSE}\\\hline
A3TGCN    & 0.3962$_{\pm0.0021}$ & \textbf{1.0416$_{\pm 0.0047}$} & 0.3401$_{\pm0.0008}$ & 0.3893$_{\pm0.0039}$ & 0.2203$_{\pm 0.0105}$ &0.2540$_{\pm0.0039}$ & 0.1373$_{\pm 0.0300}$    &  0.0722$_{\pm 0.0053}$\\
DRCNN     & 0.3499$_{\pm0.0006}$ & 1.0686$_{\pm 0.0012}$ & \first{0.1218$_{\pm0.0013}$}& \textbf{0.0960$_{\pm0.0017}$} & \first{0.0569$_{\pm 0.0004}$} & \textbf{0.0398$_{\pm0.0002}$} & \first{0.0153$_{\pm0.0002}$} & \textbf{0.0018$_{\pm4\cdot 10^{-5}}$} \\
GCRN-GRU  & \first{0.3481$_{\pm0.0008}$} &1.0534$_{\pm 0.0062}$ & 0.1219$_{\pm0.0007}$ &0.0973$_{\pm0.0009}$ & 0.0571$_{\pm 0.0005}$ &0.0404$_{\pm0.0007}$ & \first{0.0153$_{\pm 0.0003}$}    &  \textbf{0.0018$_{\pm 0.0001}$} \\
GCRN-LSTM &  0.3486$_{\pm 0.0026}$ & 1.0451$_{\pm 0.0099}$ & 0.1235$_{\pm0.0009}$&0.0985$_{\pm0.0004}$ & 0.0593$_{\pm 0.0004}$ &0.0436$_{\pm0.0001}$ & \first{0.0153$_{\pm 0.0006}$}    &  \textbf{0.0018$_{\pm 0.0001}$} \\
TGCN      & 0.4024$_{\pm 0.0022}$ & 1.0678$_{\pm 0.0049}$ & 0.3422$_{\pm 0.0046}$ & 0.3891$_{\pm0.0058}$ & 0.2109$_{\pm 0.0039}$ & 0.2466$_{\pm0.0033}$ & 0.1375$_{\pm 0.0299}$    &  0.0724$_{\pm 0.0052}$\\\specialrule{.15em}{.05em}{.05em} 

\end{tabular}
\end{table*}
Overall, DCRNN and GCRN-GRU achieve the better performance on the selected tasks. Interestingly, they both rely on Chebyshev spectral convolution and GRU, but with different architectural structure. Indeed, DCRNN employs a stacked architecture, while GCRN-GRU embeds the DGN into the RNN, enabling a combined modeling of the temporal and spatial information. This result shows that there is not a superior architectural design, in these tasks. However, it seems relevant to include a bigger neighborhood in the computation (\eg by exploiting a larger Chebishev polynomial filter size). Indeed, even if A3TGCN employs an attention mechanism to capture more global information, it is not enough to achieve comparable performance to DCRNN or GCRN-based approaches. Nevertheless, it is noteworthy that the superior performance of these approaches comes at the expense of computational speed, as it is shown in 
the Supplementary materials (Appendix~\ref{app:c}).

\subsection{D-TDG benchmark}
In the setting of general D-TDGs (\ie both nodes' state and topology may evolve over time), we consider the following datasets:
\begin{itemize}
    \item \textbf{Twitter Tennis}~\cite{twitter_tennis}: a mention graph in which nodes are Twitter accounts and their labels encode the number of mentions between them; 
    \item \textbf{Elliptic}~\cite{elliptic}: a network of bitcoin transactions, wherein a node represents a transaction and an edge indicate the payment flow. Node are also mapped to real entities belonging to licit categories (\eg exchanges, wallet providers, miners, licit services) versus illicit ones (\eg scams, malware, terrorist organizations, ransomware, Ponzi schemes);
    \item \textbf{AS-773}~\cite{as733}: the communication network of who-talks-to-whom defined in a timespan of almost 26 months from the BGP (Border Gateway Protocol) logs;
   \item \textbf{Bitcoin-$\alpha$}~\cite{bc-otc, bc-otc2}: a who-trusts-whom network of bitcoin users trading on the platform http://www.bitcoin-alpha.com. To convert this graph into a succession of snapshots, we adopted the same daily aggregation strategy as in \cite{egcn}.
\end{itemize}

We use the first two datasets to run node-level tasks. Specifically, similarly to the case of spatio-temporal setting, in Twitter tennis we perform temporal node regression, while in the Elliptic dataset \textit{temporal node classification}. Therefore, we predict the class associated to the nodes of the snapshot at time $t$ given the past graph history, $[\mathcal{G}_i]_{i=1}^t$.
We employ the last two datasets for \textit{link prediction} task, \ie to predict the future topology of the graph given its past history.

In this benchmark we evaluate three different classes of architectures (\ie stacked, integrated and meta) and we show the potential of randomized networks in the tradeoff between performance and complexity. Thus, we consider five DGNs for our experiments: DynGESN~\cite{dyngesn}, EvolveGCN-H~\cite{egcn}, EvolveGCN-O~\cite{egcn}, GCLSTM~\cite{GC-LSTM}, LRGCN~\cite{LRGCN}.


We performed hyper-parameter tuning via grid search, optimizing the Mean Absolute Error (MAE) in the case of node regression, Area Under the ROC curve (AUROC) 
in the case of link prediction, and balanced accuracy for node classification. We considered the same experimental setting, split strategy, and architectural choice as for the spatio-temporal graphs. In the case of link prediction, we perform negative sampling by randomly sampling non-occurring links from the next future snapshots. We note that in the case of DynGESN, the model employs fixed and randomized weights and only the final readout is trained.
We report in Table~\ref{tab:dtdg_grid} the grid of hyper-parameters exploited for this experiment. 
\renewcommand{\arraystretch}{1.4}
\begin{table}[ht]
\centering
\caption{The grid of hyper-parameters employed during model selection for the D-TDG tasks. The ``$*$" value refer only to LRGCN model, while ``$\diamond$" to DynGESN. \label{tab:dtdg_grid}}
\scriptsize
\begin{tabular}{ll}
\specialrule{.15em}{.05em}{.05em} 
 & \textbf{Configs}\\\hline
learning rate & $10^{-2}$, $10^{-3}$, $10^{-4}$\\
weight decay & $10^{-3}$, $10^{-4}$\\
embedding dim & 8, 16, 32\\
$\sigma$ & ReLU\\
Chebishev poly. filter size & 1, 2, 3\\
normalization scheme for $\mathbf{L}$ & \makecell[l]{$\mathbf{L} = \mathbf{D} - \mathbf{A}$,\\$\mathbf{L} = \mathbf{I} - \mathbf{D}^{-1/2} \mathbf{A} \mathbf{D}^{-1/2}$,\\$\mathbf{L} = \mathbf{I} - \mathbf{D}^{-1} \mathbf{A}$} \\
\makecell[l]{n. bases in the basis-decomposition\\regularization scheme$^*$} & None, 10, 30\\
leakage (\ie $\gamma$)$^\diamond$  & 0.1, 0.5, 0.9\\
random weight init. value$^\diamond$  & 0.1, 0.5, 0.9\\
\specialrule{.15em}{.05em}{.05em} 
\end{tabular}
\end{table}

\paragraph*{Results}
Table~\ref{tab:dtdg} shows the results on general D-TDGs. 
Differently than the spatio-temporal setting, different tasks benefit from different architectures. Indeed, integrating topology's changes (such as in GCLSTM and LRGCN) is more effective in link prediction tasks, while evolving the parameters of the DGN is more beneficial for node-level tasks, since it is more difficult to change the parameters of a static DGN to predict the topological evolution of the system. Notably, DynGESN achieves comparable results by exploiting only few trainable parameters and reduced computational overhead (see Supplementary materials, Appendix~\ref{app:c}), showing an advantageous tradeoff between performance and complexity. This makes it an ideal choice when the computational resources are limited. 
\renewcommand{\arraystretch}{1.4}
\begin{table*}[ht]
\centering
\caption{Mean test scores and std of DGNs for general D-TDGs averaged over 5 random weight initializations. For MAE and MSE scores, lower values corresponds to better performances, while for Balacend Acc, AUROC 
and F1 the higher values are better. In \first{red} is reported the optimized metric. \label{tab:dtdg}}
\scriptsize
\begin{tabular}{l  c c | c c | c c | c c}
\specialrule{.15em}{.05em}{.05em} 
& \multicolumn{4}{c}{\textbf{Node-level tasks}} & \multicolumn{4}{c}{\textbf{Link-level tasks}}\\
\cline{2-5}\cline{6-9}
     
     & \multicolumn{2}{c}{\textbf{Twitter tennis}}                                & \multicolumn{2}{c|}{\textbf{Elliptic}}                     & \multicolumn{2}{c}{\textbf{AS-773}}                         & \multicolumn{2}{c}{\textbf{Bitcoin $\alpha$}}\\
    \textbf{Model}               & \textbf{MAE}                 & \textbf{MSE}                  &  \textbf{Balanced Acc}      & \textbf{F1}                 & \textbf{AUROC}             & \textbf{F1}                    & \textbf{AUROC}               & \textbf{F1}                  \\\hline
     DynGESN       & 0.1944$_{\pm0.0056}$         & 0.3708$_{\pm0.0411}$          &  \first{50.56$_{\pm 1.10}$} & 79.2$_{\pm 19.62}$          & 95.34$_{\pm 0.04}$         & 79.83$_{\pm 5.27}$             & 97.68$_{\pm 0.12}$           & 69.98$_{\pm 1.57}$           \\
     EvolveGCN-H   & 0.1735$_{\pm0.0007}$         & 0.2858$_{\pm0.0074}$          &  49.52$_{\pm 1.55}$         & \textbf{92.54$_{\pm 8.39}$} & 59.52$_{\pm 17.53}$        & 39.85$_{\pm 34.24}$            & 51.35$_{\pm 2.88}$           & 29.55$_{\pm 30.58}$          \\
     EvolveGCN-O   & 0.1749$_{\pm0.0007}$         & 0.3020$_{\pm0.0111}$          &  49.23$_{\pm 1.03}$         & 90.80$_{\pm 12.67}$         & 58.90$_{\pm 17.80}$        & 29.99$_{\pm 37.10}$            & 51.42$_{\pm 2.84}$           & 31.74$_{\pm 29.98}$          \\
     GCLSTM        & \first{0.1686$_{\pm0.0015}$} & 0.2588$_{\pm0.0049}$          &  48.20$_{\pm 1.80}$         & 70.84$_{\pm 30.01}$         & \first{96.35$_{\pm 0.01}$} & \textbf{91.22$_{\pm 0.13}$}    & 97.75$_{\pm 0.17}$           & 91.22$_{\pm 1.38}$           \\
     LRGCN         & 0.1693$_{\pm0.0014}$         & \textbf{0.2507$_{\pm0.0057}$} &  47.84$_{\pm 3.37}$         & 65.69$_{\pm 20.21}$         & 94.77$_{\pm 0.23}$         & 89.59$_{\pm 0.33}$             & \first{98.05$_{\pm 0.03}$}   & \textbf{91.33$_{\pm 0.08}$}  \\
     \specialrule{.15em}{.05em}{.05em} 
     \end{tabular}
     \end{table*}

\subsection{C-TDG benchmark}
In the continuous scenario, we perform our experiment leveraging three datasets:
\begin{itemize}
    \item \textbf{Wikipedia}~\cite{jodie}: one month of interactions (\ie 157,474 interactions) between user and Wikipedia pages. Specifically, it corresponds to the edits made by 8,227 users on the 1,000 most edited Wikipedia pages;
    \item \textbf{Reddit}~\cite{jodie}: one month of posts (\ie interactions) made by 10,000 most active users on 1,000 most active subreddits, resulting in a total of 672,447 interactions;
    \item \textbf{LastFM}~\cite{jodie}: one month of who-listens-to-which song information. The dataset consists of 1000 users and the 1000 most listened songs, resulting in 1,293,103 interactions.
\end{itemize}

For all the datasets we considered the task of \textit{future link prediction}, thus, predicting if a link between two nodes $u$ and $v$ exists at a future time $t$ given the history of past events.

For our experimental purposes, we consider the following DGNs: DyRep~\cite{dyrep}, JODIE~\cite{jodie}, TGAT~\cite{TGAT}, and TGN~\cite{tgn_rossi2020}. These methods allow us to evaluate the sequential encoding of spatial and temporal information as well as integrated architectures. Moreover, they allow assessing the contribution of attention mechanism, embedding trajectories, and memory components. We consider as additional baseline EdgeBank~\cite{edgebank} with the aim of showing the performance of a simple heuristic. EdgeBank is a method that merely stores previously observed interactions (without any learning), and then predicts stored links as positive.

We performed hyper-parameter tuning via grid search, optimizing the AUROC score. We considered the same experimental setting and split strategy as previous experiments. We perform negative sampling by randomly sampling non-occurring links in the graph, as follows: (1) during training we sample negative destinations only from nodes that appear in the training set, (2) during validation we sample them from nodes that appear in training set or validation set and (3) during testing we sample them from the entire node set.

We report in Table~\ref{tab:ctdg_grid} the grid of hyper-parameters exploited for this experiment. 
\renewcommand{\arraystretch}{1.4}
\begin{table}[ht]
\centering
\caption{The grid of hyper-parameters employed during model selection for the C-TDG tasks. \label{tab:ctdg_grid}}
\scriptsize
\begin{tabular}{ll}
\specialrule{.15em}{.05em}{.05em} 
 & \textbf{Configs}\\\hline
learning rate & $10^{-3}$,$10^{-4}$\\
weight decay & $10^{-4}$,$10^{-5}$\\
n. DGN layers & 1, 3 \\
embedding dim & 32, 64, 96\\
DGN dim & emb dim, emb dim / 2 \\
$\sigma$ & tanh\\
Neighborhood sampler size & 5\\
\specialrule{.15em}{.05em}{.05em} 
\end{tabular}
\end{table}

\paragraph*{Results}
We report the results of the C-TDG experiments in Table~\ref{tab:ctdg}. 
\renewcommand{\arraystretch}{1.4}
\begin{table*}[ht]
\centering
\caption{Mean test scores and std of DGNs for C-TDGs averaged over 5 random weight initializations. The higher, the better. The models are trained to maximize the AUROC score.}\label{tab:ctdg}
\scriptsize
\begin{tabular}{lcc|cc|cc}
\specialrule{.15em}{.05em}{.05em} 
    & \multicolumn{2}{c}{\textbf{Wikipedia}} & \multicolumn{2}{c}{\textbf{Reddit}} & \multicolumn{2}{c}{\textbf{LastFM}}  \\
    \textbf{Model}              & \textbf{AUROC}     & \textbf{F1}         & \textbf{AUROC}     & \textbf{F1} & \textbf{AUROC}     & \textbf{F1} \\\hline
    EdgeBank        & 91.82             & \textbf{91.09}              & 96.42             & \textbf{96.29}            &   \first{94.72}       & \textbf{94.43}\\
    DyRep           & 89.72$_{\pm0.59}$ & 79.02$_{\pm0.91}$  & 97.69$_{\pm0.04}$ & 92.12$_{\pm0.13}$ & 78.41$_{\pm0.50}$ & \textbf{71.80$_{\pm0.92}$}\\
    JODIE           & 94.94$_{\pm0.48}$ & 87.52$_{\pm0.39}$  & 96.72$_{\pm0.21}$ & 89.97$_{\pm0.66}$ & 69.32$_{\pm1.33}$ & 63.95$_{\pm2.64}$\\
    TGAT            & 95.54$_{\pm0.22}$ & 88.11$_{\pm0.45}$  & 98.41$_{\pm0.01}$ & 93.58$_{\pm0.05}$ & \first{81.97$_{\pm0.08}$} & 70.96$_{\pm0.24}$\\
    TGN             & \first{97.07$_{\pm0.15}$} & \textbf{90.49$_{\pm0.24}$}  & \first{98.66$_{\pm0.04}$}    & \textbf{94.20$_{\pm0.15}$}   & 79.84$_{\pm1.58}$ & 71.09$_{\pm2.36}$\\
    \specialrule{.15em}{.05em}{.05em} 
\end{tabular}
\end{table*}
Overall, TGN generally outperforms all the other methods, showing consistent improvements over DyRep and JODIE. This result shows how the spatial information is fundamental for the effective resolution of the tasks. Indeed, an advantage of TGAT and TGN is that they can exploit bigger neighborhoods with respect to DyRep, which uses the information coming from one-hop distance, and JODIE, which only encode the source and destination information. Despite these results, we observe that the temporal information is still extremely relevant to achieve good performance. In fact, the EdgeBank baseline is able to exceed 91\% AUROC score by only looking at the graph's history. 
This is even more evident in the LastFM task, which, as observed in \cite{edgebank}, contains more reoccurring edges with respect to Wikipedia and Reddit. Consequently, such a task is comparatively easier to solve by solely exploiting these temporal patterns. Considering that EdgeBank's performance is directly correlated to the number of memorized edges, in this task, it is able to outperform all the other methods.
Lastly, it is worth mentioning that the enhanced performance of TGN and TGAT is accompanied by a trade-off in computational speed, as illustrated in 
the Supplementary materials (Appendix~\ref{app:c}). 

\section{Conclusions}\label{sec:conclusions}
Despite the field of representation learning for (static) graphs is now a consolidated and vibrant research area, there is still a strong demand for work in the domain of \textit{dynamic} graphs. 

In light of this, in this paper we proposed, at first, a survey that focuses on recent representation learning techniques for dynamic graphs under a uniform formalism consolidated from existing literature. Second, we provide the research community with a fair performance comparison among the most popular  methods of the three families of dynamic graph problems, by leveraging a reproducible experimental environment. We believe that this work will help fostering the research in the domain of dynamic graphs by providing a clear picture of the current development status and a good baseline to test new architectures and approaches. 

In order to further improve the maturity of representation learning for dynamic graphs, we believe that certain aspects should be deepened. 
A future interesting direction, in this sense, is to extend the work that has been done for \textit{heterophilic} (static) graphs~\cite{geom-gcn, heterophily_results2, cavallo2023gcnh} to the temporal domain. This will require addressing the problem of generating information-rich node representations when neighboring nodes tend to belong to different classes. A similar challenge is the one of \textit{heterogeneus} 
graphs~\cite{heterogeneous, heterogeneous2}, which contain different types of nodes and links. In this scenario, new architectures should learn the semantic-level information coming from node and edge types, in addition to topological and label information. While these are interesting future directions, we observe that there are compelling challenges that need addressing and that relate to studying, in the temporal domain, aspects such as robustness to adversarial attacks~\cite{Maddalena2022, adversarial_graph}, over-smoothing~\cite{over-smoothing} (a phenomenon where all node features become almost indistinguishable after few embedding updates), over-squashing~\cite{bottleneck, digiovanni2023oversquashing} (phenomenon preventing DGNs to propagate and preserve long-term dependencies between nodes), TDGs' expressive power~\cite{GIN, GNNBook-ch5-li}, and whole graph learning~\cite{multigraph}.

\section*{Acknowledgements}
This work has been partially supported by
EU NextGenerationEU programme under the funding schemes PNRR-PE-AI FAIR (Future Artificial Intelligence Research) and by the EU H2020 TAILOR project, GA n. 952215. The authors would like to thank Federico Errica,
NEC Laboratories Europe GmbH, for the insightful discussions throughout the development of this work.

\bibliographystyle{IEEEtranN}


\vspace{-1cm}
\begin{IEEEbiography}[{\includegraphics[width=1in,height=1.25in,clip,keepaspectratio]{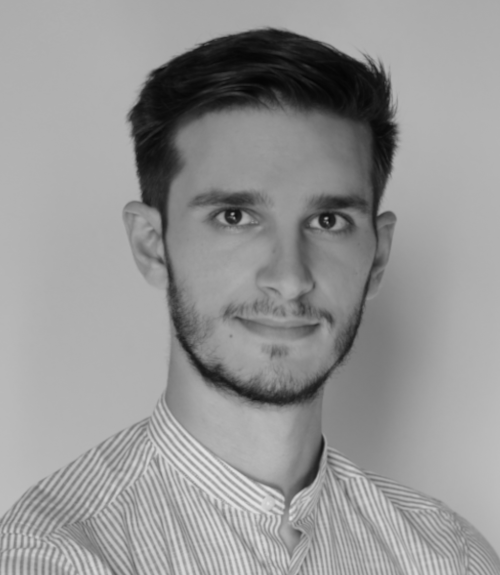}}]{Alessio Gravina} is a Ph.D. Student in Computer Science at University of Pisa. He received his B.Sc./M.Sc. in Computer Science from University of Pisa in 2018 and 2020, respectively. He was a visiting researcher at Huawei Research Center, Munich in 2023, at the Swiss AI Lab IDSIA in 2022 and at Stanford University in 2019, while, in 2018, he won the Fujistu AI-NLP Challenge. 
His interests are related to the area of machine learning for graphs and deep learning.
\end{IEEEbiography}
\vspace{-1.5cm}
\begin{IEEEbiography}[{\includegraphics[width=1in,height=1.25in,clip,keepaspectratio]{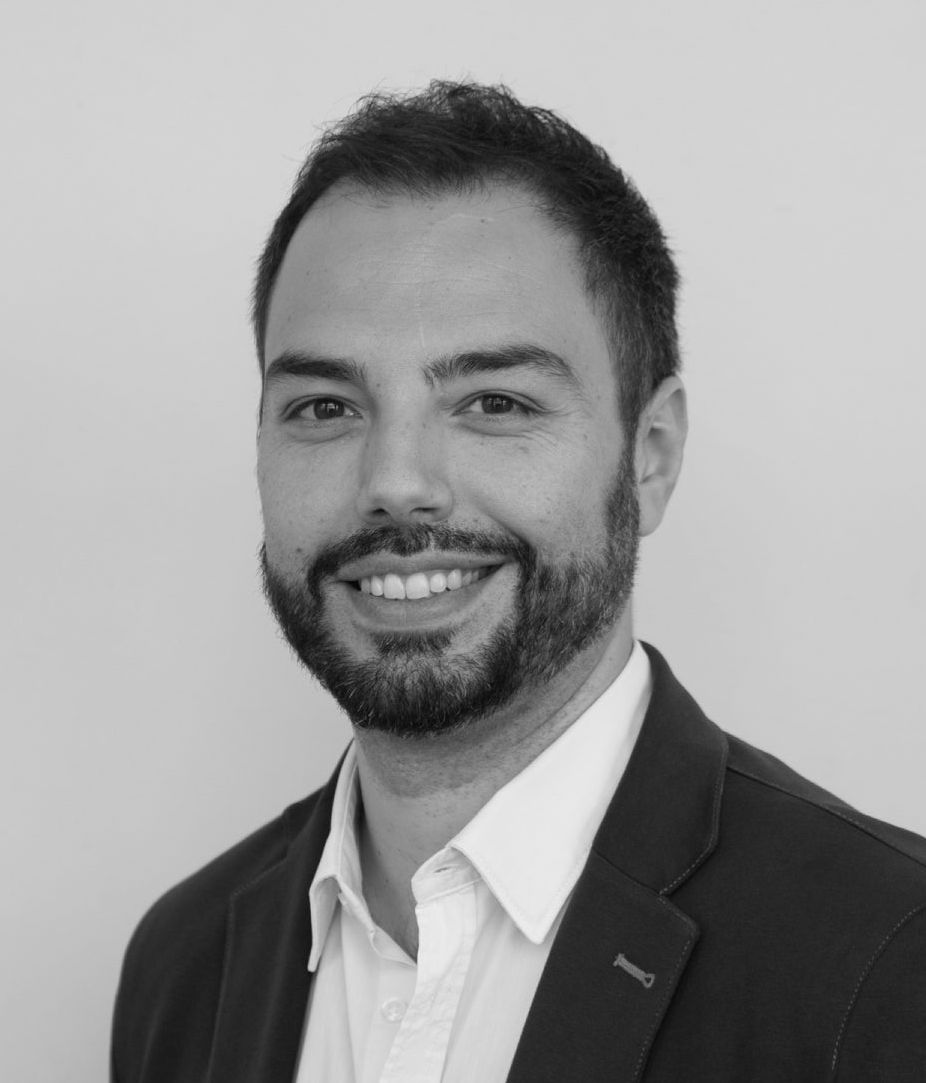}}]{Davide Bacciu} (S'06--M'09--SM'18) has a Ph.D.~in Computer Science and Engineering from IMT Lucca. He is  Full Professor at the Computer Science Department, University of Pisa, where he heads the Pervasive AI Lab. His research interests include machine learning for structured data, Bayesian learning, deep learning, reservoir computing, distributed and embedded learning systems. He is the  Chair of the IEEE NNTC and the founder of the IEEE CIS Task Force on Learning for Graphs. He is a Senior Editor of the IEEE TNNLS.
\end{IEEEbiography}


\newpage
\onecolumn
\begin{center}
\textbf{\Huge Supplementary Material of \\ ``Deep learning for dynamic graphs: models and benchmarks"}
\end{center}

\appendix



\subsection{Datasets, Models, and previous studies}\label{app:a}
In Table~\ref{tab:datasets_survey} we provide the community with a selection of datasets useful for benchmarking future works. In Table~\ref{tab:methods_wrt_events} we report an overview of the examined models with respect to the specific changes in the graph structure that each model was designed to address, \ie node/edge additions/deletions. We observe that each method is designed to address changes in node/edge features. Furthermore, the methods developed for D-TDGs that are not specifically designed to address changes in the node set can still be applied to tasks involving an evolving node set by treating nodes not in the current snapshot as isolated entities. Lastly, in Table~\ref{tab:comparison_survey} we show a comparative analysis with previous benchmarking studies and surveys, assessing the provision of datasets and benchmarks and delineating the types of analyzed graphs.

\renewcommand{\arraystretch}{1.2}
\begin{table*}[hb]
\tiny
\centering
\caption{A selected list of datasets used in dynamic graphs representation learning field. The ``$\mathcal{C}$" in the type column means C-TDG, ``$\mathcal{D}$" corresponds to D-TDG, and ``$\mathcal{ST}$" to spatio-temporal graph.\label{tab:datasets_survey}}
\begin{tabular}{l|c|c|c|c|c|c|l}
\toprule
\textbf{Name} &
  \textbf{\#Nodes} &
  \textbf{\#Edges} &
  \textbf{Seq. len.} &
  \makecell{\textbf{Snapshot sizes}\\\textbf{(nodes/edges)}} & \textbf{Granularity}& \textbf{Type}&
  \textbf{Link}\\\midrule
  
Autonomous systems  & 
  7,716 & 13,895 & 733 & 103-6,474 / 243-13,233 & daily & $\mathcal{D}$ & {\tiny\url{http://snap.stanford.edu/data/as-733.html}}  \\ \midrule
  
Bitcoin-$\alpha$ &
  3,783 &  24,186 &  24,186 & $-$  & seconds & $\mathcal{C}$ & {\tiny\url{http://snap.stanford.edu/data/soc-sign-bitcoin-alpha.html}} \\ \midrule
  
Bitcoin-OTC &
  5,881 &  35,592 &  35,592 & $-$ & seconds & $\mathcal{C}$ & {\tiny\url{http://snap.stanford.edu/data/soc-sign-bitcoin-otc.html}} \\ \midrule

CONTACT &
  274 &  2,712 &  28,244 & $-$  & $-$ & $\mathcal{C}$ &{\tiny\url{https://networkrepository.com/ia-contact.php}}\\ \midrule
  
ENRON &
  151 & 2,227 & 50,572 & $-$ & unix timestamp & $\mathcal{C}$ &{\tiny\url{https://networkrepository.com/ia-enron-employees.php}} \\ \midrule
  
Elliptic &
  203,769 &  234,355 &  49 & 1,552-12,856 / 1,168-9,164 &  49 steps & $\mathcal{D}$ &{\tiny\url{https://www.kaggle.com/ellipticco/elliptic-data-set}} \\ \midrule
  
FB-Forum &  
  899 & 7,089 & 33,700 & $-$  &  $-$ &  $\mathcal{C}$ &{\tiny\url{https://networkrepository.com/fb-forum.php}} \\ \midrule
  
FB-Covid19 &
   \makecell{152(ENG)\\104(ITA)\\95(FRA)\\53(ESP)} & \makecell{2,347(ENG)\\771(ITA)\\864(FRA)\\145(ESP)} &
   \makecell{61(ENG)\\105(ITA)\\78(FRA)\\122(ESP)}& 
   \makecell{152 / 2,347(ENG)\\104 / 771(ITA)\\95 / 864(FRA)\\53 / 145(ESP)} &  daily &  $\mathcal{D}$ &{\tiny\url{https://github.com/geopanag/pandemic_tgnn}} \\ \midrule
   
Github &
  284 & 1,420 & 20,726 & $-$ &  $-$ & $\mathcal{C}$ & {\tiny\url{https://github.com/uoguelph-mlrg/LDG}} \\ \midrule
  
HEP-TH&
   27,770 & 352,807 & 3487 & 1-650 / 0-688 &  montly &  $\mathcal{D}$ &{\tiny\url{https://snap.stanford.edu/data/cit-HepTh.html}} \\ \midrule

HYPERTEXT09 &
  113 & 2,498 & 20,819 & $-$ &  seconds & $\mathcal{C}$ &{\tiny\url{https://networkrepository.com/ia-contacts-hypertext2009.php}} \\ \midrule
  
IA-Email-EU &
  986 & 24,929 &  332,334 & $-$ &  seconds & $\mathcal{C}$ & {\tiny\url{https://snap.stanford.edu/data/email-Eu-core-temporal.html}}\\ \midrule
  
LastFM & 
    2,000 & 154,993 & 1,293,103 & $-$ &  unix timestamp & $\mathcal{C}$ & {\tiny\url{http://snap.stanford.edu/jodie/lastfm.csv}} \\ \midrule

Los-loop &
   207  & 2,833 & 2,017 & 207 / 2,833 &  5 mins & $\mathcal{ST}$ &{\tiny\url{https://github.com/lehaifeng/T-GCN/tree/master/data}}\\ \midrule
   
METR-LA &
  207 &  1,515 &  34,272 &  207 / 1515 & 5 mins & $\mathcal{ST}$ &{\tiny\url{https://github.com/liyaguang/DCRNN}}\\ \midrule

Montevideo & 675 & 690 & 740 & 675 / 690 & hourly & $\mathcal{ST}$ &{\tiny\url{https://pytorch-geometric-temporal.readthedocs.io/en/latest/modules/dataset.html}}\\\midrule
    
MOOC &
   7,144 & 411,749 & 178,443 & $-$ & unix timestamp & $\mathcal{C}$ &{\tiny\url{http://snap.stanford.edu/data/act-mooc.html}} \\ \midrule

PeMS03 & 358 & 442 & 26208 & 358 / 442 & 5 mins & $\mathcal{ST}$ &{\tiny\url{https://torch-spatiotemporal.readthedocs.io/en/latest/modules/datasets.html}}\\ \midrule

PeMS04 & 307 & 209 & 16992 & 307 / 209 &5 mins & $\mathcal{ST}$ &{\tiny\url{https://torch-spatiotemporal.readthedocs.io/en/latest/modules/datasets.html}}\\ \midrule

PeMS07 & 883 & 790 & 28225 & 883 / 790 & 5 mins & $\mathcal{ST}$ &{\tiny\url{https://torch-spatiotemporal.readthedocs.io/en/latest/modules/datasets.html}}\\ \midrule

PeMS08 & 170 & 137 & 17856 & 170 / 137 & 5 mins & $\mathcal{ST}$ &{\tiny\url{https://torch-spatiotemporal.readthedocs.io/en/latest/modules/datasets.html}}\\ \midrule

PeMSBay &
    325 & 2,369 &  52,116 & 325 / 2,369 & 5 mins & $\mathcal{ST}$ &{\tiny\url{https://github.com/liyaguang/DCRNN}} \\ \midrule
    
PeMSD7 &
  228 &  19,118 &  1,989 &  228 / 19,118 &5 mins & $\mathcal{ST}$ &{\tiny\url{https://github.com/hazdzz/STGCN/tree/main/data/pemsd7-m}} \\ \midrule

RADOSLAW & 
    167 &  5,509 &   82,927 & $-$ & seconds &$\mathcal{C}$ &{\tiny\url{https://networkrepository.com/ia-radoslaw-email.php}}\\ \midrule
    
Reddit & 
    11,000 & 78,516 & 672,447 & $-$ & unix timestamp &$\mathcal{C}$ &{\tiny\url{http://snap.stanford.edu/jodie/reddit.csv}} \\ \midrule
  
Reddit Hyperlink Network &
  55,863 & 339,643 & 858,490 & $-$ & seconds &$\mathcal{C}$ &{\tiny\url{http://snap.stanford.edu/data/soc-RedditHyperlinks.html}}\\ \midrule
  
SBM-synthetic &
  1,000 & 130,415 & 50 & 1000 / 93,835-105,358 & 50 steps & $\mathcal{D}$ &{\tiny\url{https://github.com/IBM/EvolveGCN/tree/master/data}}\\ \midrule

SOC-Wiki-Elec &
  7,118 &  103,673 & 107,071 & $-$ & $-$ &$\mathcal{C}$ & {\tiny\url{https://networkrepository.com/soc-wiki-elec.php}} \\ \midrule

SZ-taxi &
   156 & 532  & 2,977  & 156 / 532 & 15 mins & $\mathcal{ST}$ & {\tiny\url{https://github.com/lehaifeng/T-GCN/tree/master/data}} \\ \midrule

Traffic &
   4,438 & 8,996 & 2,160 & 4,438 / 8,996 & hourly & $\mathcal{ST}$ &{\tiny\url{https://github.com/chocolates/Predicting-Path-Failure-In-Time-Evolving-Graphs}} \\ \midrule

Twitter-Tennis & 
    1000 & 40,839 & 120 & 1000 / 41-936 & hourly & $\mathcal{D}$ & {\tiny\url{https://pytorch-geometric-temporal.readthedocs.io/en/latest/modules/dataset.html}}\\ \midrule

UCI messages &
  1,899 & 20,296 & 59,835 & $-$ & unix timestamp &$\mathcal{C}$ &{\tiny\url{https://snap.stanford.edu/data/CollegeMsg.html}} \\ \midrule

Wikipedia &
   9,227 & 18,257 & 157,474 & $-$ & unix timestamp &$\mathcal{C}$ &{\tiny\url{http://snap.stanford.edu/jodie/wikipedia.csv}}\\ 

\bottomrule

\end{tabular}

\end{table*}
\renewcommand{\arraystretch}{1.4}
\begin{table*}[ht]
\centering
\scriptsize
\caption{An overview of the examined models and the specific changes in the graph structure that each model was designed to address, \ie node additions/deletions and edge addition/deletions.
The ``$\mathcal{C}$" in the type column means C-TDG, ``$\mathcal{D}$" corresponds to D-TDG, and ``$\mathcal{ST}$" to spatio-temporal graph.\label{tab:methods_wrt_events}}

\begin{tabular}{l|c|c|cc|cc}
\specialrule{.15em}{.05em}{.05em} 
               \multicolumn{3}{c}{}                   & \multicolumn{2}{c}{\textbf{Node}}     & \multicolumn{2}{c}{\textbf{Edge}}\\
\textbf{Model}  & \textbf{Cit.}             & \textbf{Type}  & \textbf{Addition} & \textbf{Deletion} & \textbf{Addition} & \textbf{Deletion}\\\midrule
A3TGCN         & \cite{a3tgcn}             & $\mathcal{ST}$       & \xmark            & \xmark            & \xmark            & \xmark           \\
ASTGCN         & \cite{astgcn}             & $\mathcal{ST}$       & \xmark            & \xmark            & \xmark            & \xmark           \\
CAW            & \cite{CAW}                & $\mathcal{C}$        & \cmark            & \cmark            & \cmark            & \cmark           \\
CTDNG          & \cite{temporal_node2vec}  & $\mathcal{C}$        & \cmark            & \cmark            & \cmark            & \cmark           \\
DCRNN          & \cite{DCRNN}              & $\mathcal{ST}$       & \xmark            & \xmark            & \xmark            & \xmark           \\
DyGrAE         & \cite{dygrae}             & $\mathcal{D}$        & \cmark            & \cmark            & \cmark            & \cmark           \\
DyRep          & \cite{dyrep}              & $\mathcal{C}$        & \cmark            & \cmark            & \cmark            & \cmark           \\
DynGEM         & \cite{dyngem}             & $\mathcal{D}$        & \cmark            & \cmark            & \cmark            & \cmark           \\
DynGESN        & \cite{dyngesn}            & $\mathcal{D}$        & \cmark            & \cmark            & \cmark            & \cmark           \\
DynGraph2Vec   & \cite{dyngraph2vec}       & $\mathcal{D}$        & \cmark            & \cmark            & \cmark            & \cmark           \\
E-GCN          & \cite{egcn}               & $\mathcal{D}$        & \cmark            & \cmark            & \cmark            & \cmark           \\
Evolve2Vec     & \cite{evolve2vec}         & $\mathcal{D}$        & \cmark            & \cmark            & \cmark            & \cmark           \\
GC-LSTM        & \cite{GC-LSTM}            & $\mathcal{D}$        & \xmark            & \xmark            & \cmark            & \cmark           \\
GCRN           & \cite{GCRN}               & $\mathcal{ST}$       & \xmark            & \xmark            & \xmark            & \xmark           \\
JODIE          & \cite{jodie}              & $\mathcal{C}$        & \cmark            & \cmark            & \cmark            & \cmark           \\
LRGCN          & \cite{LRGCN}              & $\mathcal{D}$        & \xmark            & \xmark            & \cmark            & \cmark           \\
MPNN-LSTM      & \cite{mpnn_lsltm}         & $\mathcal{D}$        & \xmark            & \xmark            & \cmark            & \cmark           \\
NeurTW        & \cite{neurtws}            & $\mathcal{C}$        & \cmark            & \cmark            & \cmark            & \cmark           \\
PINT           & \cite{pint}               & $\mathcal{C}$        & \cmark            & \cmark            & \cmark            & \cmark           \\
ROLAND         & \cite{roland}             & $\mathcal{D}$        & \cmark            & \cmark            & \cmark            & \cmark           \\
SGP            & \cite{cini2023scalable}   & $\mathcal{D}$        & \xmark            & \xmark            & \cmark            & \cmark           \\
STGCN          & \cite{STGCN}              & $\mathcal{ST}$       & \xmark            & \xmark            & \xmark            & \xmark           \\
StreamGNN      & \cite{streamgnn}          & $\mathcal{C}$        & \cmark            & \cmark            & \cmark            & \cmark           \\
T-GCN          & \cite{T-GCN}              & $\mathcal{ST}$       & \xmark            & \xmark            & \xmark            & \xmark           \\
TGAT           & \cite{TGAT}               & $\mathcal{C}$        & \cmark            & \cmark            & \cmark            & \cmark           \\
TGN            & \cite{tgn_rossi2020}      & $\mathcal{C}$        & \cmark            & \cmark            & \cmark            & \cmark           \\
\specialrule{.15em}{.05em}{.05em} 

\end{tabular}
\end{table*}
\begin{table*}[ht]
    \centering
    \caption{Comparative analysis with previous benchmarking studies and surveys, assessing the provision of datasets and benchmarks and delineating the types of analyzed graphs. The ``$\mathcal{C}$" in the columns means C-TDG, ``$\mathcal{D}$" corresponds to D-TDG, and ``$\mathcal{ST}$" to spatio-temporal graph.\label{tab:comparison_survey}\\}
    \scriptsize
\begin{adjustbox}{center}
\begin{tabular}{l|cccc|c|cc|ccc}
    \toprule
    & \multicolumn{4}{c|}{Survey}             & \multirow{2}{*}{\makecell{Year of the last\\surveyed method}}                & \multicolumn{2}{c|}{Datasets} & \multicolumn{3}{c}{\makecell{Dynamic graph\\benchmark}}\\
    & Static & $\mathcal{ST}$ & $\mathcal{D}$ & $\mathcal{C}$ & & static & dynamic              & $\mathcal{ST}$ & $\mathcal{D}$ & $\mathcal{C}$\\

    \midrule
    \citet{hamilton} & \cmark & \xmark & \xmark & \xmark & 2017 & \xmark & \xmark & \xmark & \xmark & \xmark \\
    \citet{BACCIU2020203}   & \cmark & \xmark & \xmark & \xmark & 2020 & \xmark & \xmark & \xmark & \xmark & \xmark \\
    \citet{GNNsurvey}       & \cmark & \cmark & \xmark & \xmark & 2019 & \cmark & \xmark & \xmark &\xmark &\xmark \\
    \citet{dynamicgraph_survey}    & \cmark & \cmark & \cmark & \cmark & 2020 & \xmark & \cmark & \xmark &\xmark & \xmark \\
    \citet{traffic_forecasting_survey}    & \cmark & \cmark & \xmark & \xmark & 2022 & \xmark & \cmark{\tiny($\mathcal{ST}$ only)} & \xmark & \xmark & \xmark \vspace{5pt}\\
    Our                 & \cmark & \cmark & \cmark & \cmark & 2023 & \xmark & \cmark & \cmark & \cmark & \cmark \\
    \bottomrule
\end{tabular}
\end{adjustbox}
\end{table*}

\subsection{Selected Hyper-parameters}\label{app:b}
To favour reproducibility of the results and transparency of our benchmarking analysis, in Table~\ref{tab:hyper_st}, Table~\ref{tab:hyper_dt}, and Table~\ref{tab:hyper_ct} we report the hyper-parameters selected for benchmarking task and model assessed in our empirical analysis.

\begin{table*}[htb]
\centering
\caption{The table of selected hyper-parameters for the spatio-temporal tasks.}\label{tab:hyper_st}
\scriptsize
\begin{tabular}{llccccc}
\specialrule{.15em}{.05em}{.05em} 
& \textbf{Model}&  \textbf{embedding dim} &    \textbf{lr} &  \textbf{weight decay} & \textbf{filter size} &    \textbf{normalization} \\
\hline
\multirow{5}{*}{\rotatebox[origin=c]{90}{\textbf{Montevideo}}} &
 A3TGCN    &              8 &  $10^{-2}$ &        $10^{-3}$ &           - &                - \\
& DCRNN     &              8 &  $10^{-2}$ &        $10^{-3}$ &         1 &                 - \\
& GCRN-GRU  &              8 &  $10^{-2}$ &        $10^{-3}$ &           3 &             $\mathbf{D}- \mathbf{A}$ \\
& GCRN-LSTM &              8 &  $10^{-2}$ &        $10^{-4}$ &           3 &             $\mathbf{D}- \mathbf{A}$ \\
& TGCN      &              8 &  $10^{-2}$ &        $10^{-3}$ &               - &             - \\
\midrule
\multirow{5}{*}{\rotatebox[origin=c]{90}{\textbf{Metr-LA}}} &
A3TGCN    &              8 &  $10^{-3}$ &        $10^{-3}$ &               - &             - \\
& DCRNN     &              8 &  $10^{-3}$ &        $10^{-3}$ &         2 &                - \\
& GCRN-GRU  &              8 &  $10^{-3}$ &        $10^{-3}$ &             3 &             $\mathbf{D}- \mathbf{A}$ \\
& GCRN-LSTM &              8 &  $10^{-3}$ &        $10^{-3}$ &             3 &           $\mathbf{I} - \mathbf{D}^{-1/2} \mathbf{A} \mathbf{D}^{-1/2}$ \\
& TGCN      &              8 &  $10^{-3}$ &        $10^{-4}$ &               - &             - \\
\midrule
\multirow{5}{*}{\rotatebox[origin=c]{90}{\textbf{PeMSBay}}} &
A3TGCN    &              8 &  $10^{-3}$ &        $10^{-4}$ &               - &             - \\
& DCRNN     &              8 &  $10^{-4}$ &        $10^{-4}$ &         2 &                 - \\
& GCRN-GRU  &              8 &  $10^{-4}$ &        $10^{-4}$ &             2 &             $\mathbf{D}- \mathbf{A}$ \\
& GCRN-LSTM &              8 &  $10^{-4}$ &        $10^{-4}$ &             3 &            $\mathbf{D}^{-1} \mathbf{A}$ \\
& TGCN      &              8 &  $10^{-4}$ &        $10^{-4}$ &               - &             - \\

\midrule
\multirow{5}{*}{\rotatebox[origin=c]{90}{\textbf{Traffic}}} &
A3TGCN    &              8 &  $10^{-2}$ &        $10^{-4}$ &               - &             - \\
& DCRNN     &        8       &  $10^{-3}$ &        $10^{-3}$ &         3 &                 - \\
& GCRN-GRU  &              8 &  $10^{-2}$ &        $10^{-4}$ &             2 &             $\mathbf{I} - \mathbf{D}^{-1/2} \mathbf{A} \mathbf{D}^{-1/2}$ \\
& GCRN-LSTM &              8 &  $10^{-2}$ &        $10^{-3}$ &             2 &            $\mathbf{I} - \mathbf{D}^{-1/2} \mathbf{A} \mathbf{D}^{-1/2}$ \\
& TGCN      &              8 &  $10^{-2}$ &        $10^{-4}$ &               - &             - \\
\specialrule{.15em}{.05em}{.05em} 
\end{tabular}
\end{table*}


\begin{table*}[ht]
    \centering
    \caption{The table of selected hyper-parameters for the D-TDG tasks.}\label{tab:hyper_dt}
    \scriptsize
\begin{tabular}{llcccccccc}
\specialrule{.15em}{.05em}{.05em} 
&     \textbf{Model}      &  \begin{tabular}{@{}c@{}}\textbf{embedding}\vspace{-3pt}\\\textbf{dim}\end{tabular} &           \textbf{lr} &  \begin{tabular}{@{}c@{}}\textbf{weight}\vspace{-3pt}\\\textbf{decay}\end{tabular} &    \textbf{n. bases} &    \textbf{K} & \textbf{normalization} & \begin{tabular}{@{}c@{}}\textbf{random weight}\vspace{-3pt}\\\textbf{init. value}\end{tabular} & \textbf{$\gamma$} \\
\hline
\multirow{5}{*}{\rotatebox[origin=c]{90}{\textbf{Twitter tennis}}} &
DynGESN      &             32 &        $10^{-2}$ &        $10^{-3}$ &             - &    - &             - &   0.9 &     0.5 \\
&EvolveGCN-H  &              8 &        $10^{-2}$ &        $10^{-3}$ &             - &    - &          $\mathbf{D} - \mathbf{A}$ &     - &       - \\
&EvolveGCN-O  &             32 &        $10^{-3}$ &        $10^{-3}$ &             - &    - &          $\mathbf{D} - \mathbf{A}$ &     - &       - \\
&GCLSTM       &             32 &        $10^{-2}$ &        $10^{-4}$ &             - &    2 &             $\mathbf{D} - \mathbf{A}$ &     - &       - \\
&LRGCN        &             32 &        $10^{-2}$ &        $10^{-4}$ &             None &    - &             - &     - &       - \\

\midrule

\multirow{5}{*}{\rotatebox[origin=c]{90}{\textbf{Elliptic}}} &
DynGESN     &              8 &         $10^{-2}$ &        $10^{-4}$ &             - &    - &             - &   0.1 &     0.9 \\
&EvolveGCN-H &             32 &         $10^{-4}$ &        $10^{-4}$ &            - &    - &             $\mathbf{I} - \mathbf{D}^{-1/2} \mathbf{A} \mathbf{D}^{-1/2}$ &     - &       - \\
&EvolveGCN-O &             16 &         $10^{-4}$ &        $10^{-3}$ &            - &    - &             $\mathbf{I} - \mathbf{D}^{-1/2} \mathbf{A} \mathbf{D}^{-1/2}$ &     - &       - \\
&GCLSTM     &              8 &         $10^{-4}$ &        $10^{-4}$ &             - &    1 &             $\mathbf{D}  - \mathbf{A}$ &     - &       - \\
&LRGCN      &              8 &         $10^{-4}$ &        $10^{-4}$ &            10 &    - &             - &     - &       - \\

\midrule
\multirow{5}{*}{\rotatebox[origin=c]{90}{\textbf{AS-733}}} &
DynGESN     &             32 &         $10^{-2}$ &        $10^{-4}$ &            - &  - &             - &   0.9 &     0.5 \\
&EvolveGCN-H &             32 &         $10^{-3}$ &        $10^{-3}$ &           - &  - &             $\mathbf{D} - \mathbf{A}$ &     - &       - \\
&EvolveGCN-O &             16 &         $10^{-3}$ &        $10^{-4}$ &           - &  - &             $\mathbf{D} - \mathbf{A}$ &     - &       - \\
&GCLSTM     &             32 &         $10^{-2}$ &        $10^{-3}$ &            - &  2 &           $\mathbf{I} - \mathbf{D}^{-1/2} \mathbf{A}\mathbf{D}^{-1/2}$ &     - &       - \\
&LRGCN      &             32 &         $10^{-3}$ &        $10^{-3}$ &           10 &  - &             - &     - &       - \\

\midrule

\multirow{5}{*}{\rotatebox[origin=c]{90}{\textbf{Bitcoin $\alpha$}}} &
DynGESN      & 32 & $10^{-2}$& $10^{-4}$ & - &    - &  - & 0.9 &     0.1\\
&EvolveGCN-H &             32 &          $10^{-4}$ &        $10^{-3}$ &                 - &    - &             $\mathbf{D} - \mathbf{A}$ &     - &       - \\
&EvolveGCN-O &             16 &          $10^{-3}$ &        $10^{-4}$ &                 - &    - &             $\mathbf{D} - \mathbf{A}$ &     - &       - \\
&GCLSTM     &             16 &          $10^{-2}$ &        $10^{-4}$ &                 - &  3 &             $\mathbf{D} - \mathbf{A}$ &     - &       - \\
&LRGCN      &             32 &          $10^{-2}$ &        $10^{-4}$ &              10 &    - &             - &     - &       - \\

\specialrule{.15em}{.05em}{.05em} 
\end{tabular}
\end{table*}


\begin{table*}[ht]
    \centering
    \caption{The table of selected hyper-parameters for the C-TDG tasks.}\label{tab:hyper_ct}
    \scriptsize
\begin{tabular}{llccccc}
\specialrule{.15em}{.05em}{.05em} 
&       \textbf{Model}    &  \textbf{embedding dim} &           \textbf{lr} &  \textbf{weight decay} &    \textbf{n. DGN layers} & \textbf{DGN dim} \\
\hline
\multirow{4}{*}{\rotatebox[origin=c]{90}{\textbf{Wikipedia}}} &
DyRep &                      96 &           $10^{-3}$ &          $10^{-4}$ &             -  & -\\
&JODIE &                      96 &           $10^{-4}$ &          $10^{-5}$ &             -  & -\\
&TGAT  &                      96 &           $10^{-3}$ &          $10^{-4}$ &           3 & 96\\
&TGN   &                      96 &           $10^{-3}$ &          $10^{-5}$ &           3 & 48\\

\midrule
\multirow{4}{*}{\rotatebox[origin=c]{90}{\textbf{Reddit}}} &
DyRep &                      96 &           $10^{-4}$ &          $10^{-4}$ &             - & -\\
&JODIE &                      96 &           $10^{-4}$ &          $10^{-5}$ &             - & -\\
&TGAT  &                      96 &           $10^{-3}$ &          $10^{-4}$ &           3 & 96\\
&TGN   &                      96 &           $10^{-3}$ &          $10^{-5}$ &           1 & 96\\

\midrule
\multirow{4}{*}{\rotatebox[origin=c]{90}{\textbf{LastFM}}} &
DyRep &                      96 &           $10^{-4}$ &          $10^{-4}$ &             - & -\\
&JODIE &                      32 &           $10^{-4}$ &          $10^{-5}$ &             - & -\\
&TGAT  &                      48 &           $10^{-4}$ &          $10^{-4}$ &           3 & 48\\
&TGN   &                      32 &           $10^{-3}$ &          $10^{-5}$ &           3 & 16\\

\specialrule{.15em}{.05em}{.05em} 

\end{tabular}
\end{table*}

\subsection{Complete results}\label{app:c}
Table~\ref{tab:dtdg_complete} and Table~\ref{tab:ctdg_complete} provide the full results for the discrete and continuous time dynamic graph tasks, including balanced accuracy as an additional metric (with respect to those in the main body of the work).  Table~\ref{tab:robustness_st}, Table~\ref{tab:robustness_dt}, and Table~\ref{tab:robustness_ct}, instead,  report the minimum and maximum standard deviation (std) of validation scores obtained by fixing individual hyper-parameter values in spatio-temporal, D-TDGs, and C-TDGs, respectively. Thus, we study the stability of training under various hyper-parameters. We observe that models typically exhibit stability across diverse hyper-parameter settings, and that weight decay is among the hyper-parameters with less influence on the final score, while the embedding dimension and learning rate are the most affecting ones. Lastly, we report the average time per forward pass on the whole dataset (measured in seconds) for each model in every task in Table~\ref{tab:time_st}, Table~\ref{tab:time_dtdg}, and Table~\ref{tab:time_ctdg}.

\renewcommand{\arraystretch}{1.4}
\begin{table*}[ht]
\centering
\caption{Mean test scores and std of DGNs for general D-TDGs averaged over 5 random weight initializations. For MAE and MSE scores, lower values corresponds to better performances, while for Balacend Acc, AUROC 
and F1 the higher values are better. In \first{red} is reported the optimized metric. 
\label{tab:dtdg_complete}}
\scriptsize
\begin{tabular}{l  c c | c c c }
\multicolumn{6}{c}{\textbf{Node-level tasks}}\\
\specialrule{.15em}{.05em}{.05em} 
& \multicolumn{2}{c}{\textbf{Twitter tennis}}                                & \multicolumn{3}{c}{\textbf{Elliptic}}\\
 \textbf{Model} & \textbf{MAE}               & \textbf{MSE}                  & \textbf{AUROC}              & \textbf{F1}                 & \textbf{Balanced Acc}  \\\hline
DynGESN       & 0.1944$_{\pm0.0056}$         & 0.3708$_{\pm0.0411}$          & \textbf{51.12$_{\pm 1.30}$} & 79.2$_{\pm 19.62}$          & \first{50.56$_{\pm 1.10}$}     \\
EvolveGCN-H   & 0.1735$_{\pm0.0007}$         & 0.2858$_{\pm0.0074}$          & 48.43$_{\pm 2.71}$          & \textbf{92.54$_{\pm 8.39}$} & 49.52$_{\pm 1.55}$     \\
EvolveGCN-O   & 0.1749$_{\pm0.0007}$         & 0.3020$_{\pm0.0111}$          & 45.11$_{\pm 1.68}$          & 90.80$_{\pm 12.67}$         & 49.23$_{\pm 1.03}$     \\
GCLSTM        & \first{0.1686$_{\pm0.0015}$} & 0.2588$_{\pm0.0049}$          & 45.77$_{\pm 1.60}$          & 70.84$_{\pm 30.01}$         & 48.20$_{\pm 1.80}$     \\
LRGCN         & 0.1693$_{\pm0.0014}$         & \textbf{0.2507$_{\pm0.0057}$} & 45.82$_{\pm 3.81}$          & 65.69$_{\pm 20.21}$         & 47.84$_{\pm 3.37}$     \\
\specialrule{.15em}{.05em}{.05em} 
\end{tabular}
\vspace{5pt}

\begin{tabular}{l  c c c | c c c }
\multicolumn{7}{c}{\textbf{Link-level tasks}}\\
\specialrule{.15em}{.05em}{.05em} 
& \multicolumn{3}{c}{\textbf{AS-773}}                                                                     & \multicolumn{3}{c}{\textbf{Bitcoin $\alpha$}}\\
      \textbf{Model}       & \textbf{AUROC}             & \textbf{F1}                  & \textbf{Balanced Acc}        & \textbf{AUROC}               & \textbf{F1}                  & \textbf{Balanced Acc} \\\hline
DynGESN      & 95.34$_{\pm 0.04}$         & 79.83$_{\pm 5.27}$           & 82.80$_{\pm 3.40}$           & 97.68$_{\pm 0.12}$           & 69.98$_{\pm 1.57}$           & 76.79$_{\pm 0.93}$ \\
EvolveGCN-H  & 59.52$_{\pm 17.53}$        & 39.85$_{\pm 34.24}$          & 53.72$_{\pm 16.79}$          & 51.35$_{\pm 2.88}$           & 29.55$_{\pm 30.58}$          & 50.69$_{\pm 1.69}$ \\
EvolveGCN-O  & 58.90$_{\pm 17.80}$        & 29.99$_{\pm 37.10}$          & 56.99$_{\pm 13.97}$          & 51.42$_{\pm 2.84}$           & 31.74$_{\pm 29.98}$          & 51.42$_{\pm 2.84}$ \\
GCLSTM       & \first{96.35$_{\pm 0.01}$} & \textbf{91.22$_{\pm 0.13}$}  & \textbf{91.11$_{\pm 0.06}$}  & 97.75$_{\pm 0.17}$           & 91.22$_{\pm 1.38}$           & 91.72$_{\pm 1.11}$ \\
LRGCN        & 94.77$_{\pm 0.23}$         & 89.59$_{\pm 0.33}$           & 89.07$_{\pm 0.34}$           & \first{98.05$_{\pm 0.03}$}   & \textbf{91.33$_{\pm 0.08}$}  & \textbf{91.89$_{\pm 0.07}$} \\
\specialrule{.15em}{.05em}{.05em} 
\end{tabular}
\end{table*}
\renewcommand{\arraystretch}{1.4}
\begin{table*}[ht]
\centering
\caption{Mean test scores and std of DGNs for C-TDGs averaged over 5 random weight initializations. The higher, the better. The models are trained to maximize the AUROC score.}\label{tab:ctdg_complete}
\scriptsize
\begin{tabular}{lccc|ccc|ccc}
\specialrule{.15em}{.05em}{.05em} 
    & \multicolumn{3}{c}{\textbf{Wikipedia}} & \multicolumn{3}{c}{\textbf{Reddit}} & \multicolumn{3}{c}{\textbf{LastFM}}  \\
   \textbf{Model}   & \textbf{AUROC}            & \textbf{F1}                 & \textbf{Acc}         & \textbf{AUROC}               & \textbf{F1}                & \textbf{Acc}      & \textbf{AUROC}            & \textbf{F1}                & \textbf{Acc}      \\\hline
    EdgeBank        & 91.82                     & \textbf{91.09}              & 91.82                & 96.42                        & \textbf{96.29}             & \textbf{96.42}             & \first{94.72}             & \textbf{94.43}             & \textbf{94.72}                 \\
    DyRep           & 89.72$_{\pm0.59}$         & 79.02$_{\pm0.91}$           & 80.46$_{\pm0.63}$    & 97.69$_{\pm0.04}$            & 92.12$_{\pm0.13}$          & 92.02$_{\pm0.19}$ & 78.41$_{\pm0.50}$         & \textbf{71.80$_{\pm0.92}$} & 68.63$_{\pm1.01}$ \\
    JODIE           & 94.94$_{\pm0.48}$         & 87.52$_{\pm0.39}$           & 87.85$_{\pm0.44}$    & 96.72$_{\pm0.21}$            & 89.97$_{\pm0.66}$          & 89.48$_{\pm0.86}$ & 69.32$_{\pm1.33}$         & 63.95$_{\pm2.64}$          & 62.08$_{\pm2.78}$  \\
    TGAT            & 95.54$_{\pm0.22}$         & 88.11$_{\pm0.45}$           & 88.58$_{\pm0.31}$    & 98.41$_{\pm0.01}$            & 93.58$_{\pm0.05}$          & 93.63$_{\pm0.04}$ & \first{81.97$_{\pm0.08}$} & 70.96$_{\pm0.24}$          & \textbf{72.64$_{\pm0.09}$}  \\
    TGN             & \first{97.07$_{\pm0.15}$} & \textbf{90.49$_{\pm0.24}$}  & \textbf{90.66$_{\pm0.22}$}    & \first{98.66$_{\pm0.04}$}    & \textbf{94.20$_{\pm0.15}$} & \textbf{94.19$_{\pm0.17}$} & 79.84$_{\pm1.58}$         & 71.09$_{\pm2.36}$          & 63.13$_{\pm7.03}$ \\
    \specialrule{.15em}{.05em}{.05em} 
\end{tabular}
\end{table*}

\begin{table*}[ht]
    \centering
    \caption{The \textbf{min}imum and \textbf{max}imum standard deviation of validation scores obtained by fixing individual hyperparameter values in spatio-temporal models. The hyperparameter names corresponding to these values are also provided for reference. ``wd" means weight decay, ``ed" embedding dimension, ``lr" learning rate, and ``fs" filter size.\label{tab:robustness_st}\\}
    \scriptsize
\begin{adjustbox}{center}
\begin{tabular}{lll|ll|ll|ll}
    \toprule
& \multicolumn{2}{c|}{\textbf{Montevideo}} & \multicolumn{2}{c|}{\textbf{MetrLA}}   & \multicolumn{2}{c|}{\textbf{PeMSBay}} & \multicolumn{2}{c}{\textbf{Traffic}}\\
\textbf{Model}&      \multicolumn{1}{c}{\textbf{min}} & \multicolumn{1}{c|}{\textbf{max}}
    &      \multicolumn{1}{c}{\textbf{min}} & \multicolumn{1}{c|}{\textbf{max}}
    &      \multicolumn{1}{c}{\textbf{min}} & \multicolumn{1}{c|}{\textbf{max}}
    &      \multicolumn{1}{c}{\textbf{min}} & \multicolumn{1}{c}{\textbf{max}}\\

\midrule
A3TGCN    & ed: 0.009 &  lr: 0.015 & wd: 0.099 &  ed: 0.127 &   ed: 0.039 &  lr: 0.081 &  ed: 0.006 &  lr: 0.009 \\
DCRNN     & ed: 0.016 &  lr: 0.025 & wd: 0.019 &  ed: 0.034 &   ed: 0.050 &  fs: 0.131 &  wd: 0.017 &  ed: 0.020 \\
GCRN-GRU  & lr: 0.038 &   K: 0.043 & wd: 0.057 &   K: 0.071 &   ed: 0.086 &   K: 0.150 &  wd: 0.040 &   K: 0.047 \\
GCRN-LSTM & wd: 0.031 &  ed: 0.039 & wd: 0.056 &  ed: 0.082 &   ed: 0.072 &   K: 0.155 &  wd: 0.024 &  lr: 0.031 \\
TGCN      & ed: 0.013 &  lr: 0.018 & wd: 0.099 &  ed: 0.120 &   ed: 0.041 &  lr: 0.087 &  ed: 0.006 &  lr: 0.009 \\
    \bottomrule
\end{tabular}
\end{adjustbox}
\end{table*}

\begin{table*}[ht]
    \centering
    \caption{The \textbf{min}imum and \textbf{max}imum standard deviation of validation scores obtained by fixing individual hyperparameter values in D-TDGs models. The hyperparameter names corresponding to these values are also provided for reference. ``wd" means weight decay, ``ed" embedding dimension, ``lr" learning rate, ``ns" normalization scheme, and ``nb" number of bases.\label{tab:robustness_dt}\\}
    \scriptsize
\begin{adjustbox}{center}
\begin{tabular}{lll|ll|ll|ll}
    \toprule
     & \multicolumn{2}{c|}{\textbf{Twitter tennis}} & \multicolumn{2}{c|}{\textbf{Elliptic}} & \multicolumn{2}{c|}{\textbf{AS-733}} & \multicolumn{2}{c}{\textbf{Bitcoin-$\alpha$}}  \\
 \textbf{Model}    &      \multicolumn{1}{c}{\textbf{min}} & \multicolumn{1}{c|}{\textbf{max}}
         &      \multicolumn{1}{c}{\textbf{min}} & \multicolumn{1}{c|}{\textbf{max}}
         &      \multicolumn{1}{c}{\textbf{min}} & \multicolumn{1}{c|}{\textbf{max}}
         &      \multicolumn{1}{c}{\textbf{min}} & \multicolumn{1}{c}{\textbf{max}}\\
 
     \midrule
 DynGESN      &  lr: 0.004 &  $\sigma$: 0.006  &   lr: 0.004 &  ed: 0.005 & ed: 0.003 &  $\sigma$: 0.005 &  $\gamma$: 0.009 &  lr: 0.052  \\
 EvolveGCN-H  &  wd: 0.011 &  lr: 0.012        &   lr: 0.003 &  ed: 0.005 & wd: 0.082 &  ed: 0.109       &  ns: 0.018       &  ed: 0.020  \\
 EvolveGCN-O  &  ed: 0.010 &  lr: 0.012        &   lr: 0.005 &  ed: 0.007 & ed: 0.095 &  wd: 0.105       &  lr: 0.022       &  wd: 0.024  \\
 GCLSTM       &  lr: 0.012 &  ed: 0.016        &   wd: 0.004 &  ed: 0.005 & K: 0.201  &  ed: 0.226       &  K: 0.025        &  lr: 0.207  \\
 LRGCN        &  nb: 0.008 &  ed: 0.01         &   wd: 0.005 &  nb: 0.005 & nb: 0.036 &  lr: 0.057       &  nb: 0.003       &  ed: 0.004  \\
    \bottomrule
\end{tabular}
\end{adjustbox}
\end{table*}

\begin{table*}[ht]
    \centering
    \caption{The \textbf{min}imum and \textbf{max}imum standard deviation of validation scores obtained by fixing individual hyperparameter values in C-TDGs models. The hyperparameter names corresponding to these values are also provided for reference. ``wd" means weight decay, ``ed" embedding dimension, ``lr" learning rate, ``gl" number of DGN layers, ``ge" DGN dimension, and ``re" readout embedding dim.\label{tab:robustness_ct}\\}
    \scriptsize
\begin{adjustbox}{center}
\begin{tabular}{lll|ll|ll}
    \toprule
&  \multicolumn{2}{c|}{\textbf{Wikipedia}} & \multicolumn{2}{c|}{\textbf{Reddit}} & \multicolumn{2}{c}{\textbf{LastFM}}  \\
\textbf{Model} &      \multicolumn{1}{c}{\textbf{min}} & \multicolumn{1}{c|}{\textbf{max}}
    &      \multicolumn{1}{c}{\textbf{min}} & \multicolumn{1}{c|}{\textbf{max}}
    &      \multicolumn{1}{c}{\textbf{min}} & \multicolumn{1}{c}{\textbf{max}}\\

\midrule
DyRep  & ed: 0.004 & wd: 0.009 &   lr: 0.002 & wd: 0.003   &   lr: 0.004 &     ed: 0.007  \\
JODIE  & lr: 0.023 & ed: 0.027 &   wd: 0.007 & ed: 0.007   &   wd: 0.004 &     ed: 0.006  \\
TGAT   & gl: 0.004 & re: 0.023 &   ge: 0.002 & re: 0.011   &   gl: 0.007 &     wd: 0.030  \\
TGN    & ed: 0.004 & wd: 0.005 &   wd: 0.004 & ge: 0.006   &   gl: 0.012 &     lr: 0.033  \\
    \bottomrule
\end{tabular}
\end{adjustbox}
\end{table*}
\renewcommand{\arraystretch}{1.4}

\begin{table*}[ht]
\centering
\caption{Average time to execute a forward pass on the whole dataset (measured in seconds) and std of the best configuration of each model in each task in the spatio-temporal setting, averaged over 5 repetitions. The evaluation were carried out on an Intel(R) Xeon(R) Gold 6240R CPU @ 2.40GHz.}\label{tab:time_st}
\scriptsize
\begin{tabular}{lcccc}
\specialrule{.15em}{.05em}{.05em} 
\textbf{Model} & \textbf{Montevideo} & \textbf{Metr-LA} & \textbf{PeMSBay}  & \textbf{Traffic}  \\\hline
A3TGCN      & \first{1.59$_{\pm0.05}$}   & 95.12$_{\pm0.53}$  & 167.05$_{\pm1.96}$ & 18.81$_{\pm1.04}$\\
DCRNN       & 3.15$_{\pm0.1}$    & 291.44$_{\pm0.76}$ & 366.34$_{\pm2.73}$ & 112.49$_{\pm11.52}$\\
GCRN-GRU    & 4.89$_{\pm0.1}$    & 216.99$_{\pm3.77}$ & 289.38$_{\pm3.38}$ & 32.44$_{\pm1.03}$\\
GCRN-LSTM   & 6.92$_{\pm0.13}$   & 313.27$_{\pm4.09}$ & 534.65$_{\pm5.85}$ & 46.08$_{\pm1.6}$\\
TGCN        & 1.64$_{\pm0.07}$   & \first{95.08$_{\pm1.43}$}  & \first{165.19$_{\pm2.69}$} & \first{17.96$_{\pm0.52}$}\\
\specialrule{.15em}{.05em}{.05em} 

\end{tabular}
\end{table*}

\begin{table*}[ht]
\centering
\caption{Average time to execute a forward pass on the whole dataset (measured in seconds) and std of the best configuration of each model in each task in the generic D-TDG, averaged over 5 repetitions. The evaluation were carried out on an Intel(R) Xeon(R) Gold 6240R CPU @ 2.40GHz.}\label{tab:time_dtdg}
\scriptsize
\begin{tabular}{lcccc}
\specialrule{.15em}{.05em}{.05em} 
\textbf{Model} & \textbf{Twitter tennis} & \textbf{Elliptic} & \textbf{AS-773}  & \textbf{Bitcoin $\alpha$}  \\\hline
DynGESN     & \first{0.04$_{\pm2\cdot10^{-3}}$}  & \first{0.03$_{\pm0.01}$}  & \first{0.53$_{\pm0.13}$} &  \first{0.15$_{\pm0.01}$}\\
EvolveGCN-H & 0.38$_{\pm0.01}$   &  21.14$_{\pm0.07}$   & 3.73$_{\pm0.18}$   & 2.33$_{\pm0.03}$\\
EvolveGCN-O & 0.11$_{\pm0.02}$  &  19.5$_{\pm0.25}$    & 2.35$_{\pm0.29}$   & 1.07$_{\pm0.03}$\\
GCLSTM      & 1.08$_{\pm0.42}$  &  1.2$_{\pm0.13}$     & 31.14$_{\pm0.83}$  & 28.37$_{\pm1.59}$\\
LRGCN       & 1.63$_{\pm0.08}$  &  3.28$_{\pm0.28}$    & 21.66$_{\pm2.65}$  & 24.77$_{\pm3.21}$\\
\specialrule{.15em}{.05em}{.05em} 

\end{tabular}
\end{table*}

\begin{table*}[ht]
\centering
\caption{Average time to execute a forward pass on the whole dataset (measured in seconds) and std of the best configuration of each model in each task in the C-TDG setting, averaged over 5 repetitions. The evaluation were carried out on an Intel(R) Xeon(R) Gold 6278C CPU @ 2.60GHz.}\label{tab:time_ctdg}
\scriptsize
\begin{tabular}{lccc}
\specialrule{.15em}{.05em}{.05em} 
 \textbf{Model}      & \textbf{Wikipedia} &  \textbf{Reddit}             &      \textbf{LastFM}\\\hline
DyRep  & 13.95$_{\pm1.05}$  &   99.11$_{\pm9.04}$  & 143.15$_{\pm12.72}$\\
JODIE  & \first{12.66$_{\pm1.98}$}  &   \first{83.27$_{\pm6.47}$}  &  \first{117.17$_{\pm6.86}$}\\
TGAT   & 36.84$_{\pm2.09}$  &    303.7$_{\pm7.8}$  & 167.65$_{\pm13.33}$\\
TGN    & 28.35$_{\pm2.14}$  & 114.73$_{\pm14.77}$  &  178.15$_{\pm7.87}$\\
\specialrule{.15em}{.05em}{.05em} 

\end{tabular}
\end{table*}



\begin{thebibliography}{93}
\providecommand{\natexlab}[1]{#1}
\providecommand{\url}[1]{#1}
\csname url@samestyle\endcsname
\providecommand{\newblock}{\relax}
\providecommand{\bibinfo}[2]{#2}
\providecommand{\BIBentrySTDinterwordspacing}{\spaceskip=0pt\relax}
\providecommand{\BIBentryALTinterwordstretchfactor}{4}
\providecommand{\BIBentryALTinterwordspacing}{\spaceskip=\fontdimen2\font plus
\BIBentryALTinterwordstretchfactor\fontdimen3\font minus \fontdimen4\font\relax}
\providecommand{\BIBforeignlanguage}[2]{{%
\expandafter\ifx\csname l@#1\endcsname\relax
\typeout{** WARNING: IEEEtranN.bst: No hyphenation pattern has been}%
\typeout{** loaded for the language `#1'. Using the pattern for}%
\typeout{** the default language instead.}%
\else
\language=\csname l@#1\endcsname
\fi
#2}}
\providecommand{\BIBdecl}{\relax}
\BIBdecl

\bibitem[Gilmer et~al.(2017)Gilmer, Schoenholz, Riley, Vinyals, and Dahl]{MPNN}
J.~Gilmer, S.~S. Schoenholz, P.~F. Riley, O.~Vinyals, and G.~E. Dahl, ``{Neural Message Passing for Quantum Chemistry},'' in \emph{Proc. of the 34th ICML}, vol.~70.\hskip 1em plus 0.5em minus 0.4em\relax JMLR, 2017, p. 1263–1272.

\bibitem[Zitnik et~al.(2018)Zitnik, Agrawal, and Leskovec]{bioinformatics}
M.~Zitnik, M.~Agrawal, and J.~Leskovec, ``{Modeling polypharmacy side effects with graph convolutional networks},'' \emph{Bioinformatics}, vol.~34, no.~13, pp. i457--i466, 2018.

\bibitem[Gravina et~al.(2022)Gravina, Wilson, Bacciu, Grimes, and Priami]{gravina_schizophrenia}
A.~Gravina, J.~L. Wilson, D.~Bacciu, K.~J. Grimes, and C.~Priami, ``{Controlling astrocyte-mediated synaptic pruning signals for schizophrenia drug repurposing with deep graph networks},'' \emph{PLoS Comput. Biol.}, vol.~18, no.~5, pp. 1--19, 2022.

\bibitem[Bacciu et~al.(2023)Bacciu, Errica, Gravina, Madeddu, Podda, and Stilo]{gravina2023Covid}
D.~Bacciu, F.~Errica, A.~Gravina, L.~Madeddu, M.~Podda, and G.~Stilo, ``{Deep Graph Networks for Drug Repurposing with Multi-Protein Targets},'' \emph{IEEE TETC}, pp. 1--14, 2023.

\bibitem[Monti et~al.(2019)Monti, Frasca, Eynard, Mannion, and Bronstein]{social_network}
F.~Monti, F.~Frasca, D.~Eynard, D.~Mannion, and M.~M. Bronstein, ``{Fake News Detection on Social Media using Geometric Deep Learning},'' \emph{arXiv preprint arXiv:1902.06673}, 2019.

\bibitem[Derrow-Pinion et~al.(2021)Derrow-Pinion, She, Wong, Lange, Hester, Perez, Nunkesser, Lee, Guo, Wiltshire, Battaglia, Gupta, Li, Xu, Sanchez-Gonzalez, Li, and Velickovic]{google_maps}
A.~Derrow-Pinion, J.~She, D.~Wong, O.~Lange, T.~Hester, L.~Perez, M.~Nunkesser, S.~Lee, X.~Guo, B.~Wiltshire, P.~W. Battaglia, V.~Gupta, A.~Li, Z.~Xu, A.~Sanchez-Gonzalez, Y.~Li, and P.~Velickovic, ``{ETA Prediction with Graph Neural Networks in Google Maps},'' in \emph{Proc. of the 30th ACM CIKM}.\hskip 1em plus 0.5em minus 0.4em\relax Association for Computing Machinery, 2021, p. 3767–3776.

\bibitem[Bacciu et~al.(2020)Bacciu, Errica, Micheli, and Podda]{BACCIU2020203}
D.~Bacciu, F.~Errica, A.~Micheli, and M.~Podda, ``{A gentle introduction to deep learning for graphs},'' \emph{Neural Networks}, vol. 129, pp. 203--221, 2020.

\bibitem[Wu et~al.(2021)Wu, Pan, Chen, Long, Zhang, and Yu]{GNNsurvey}
Z.~Wu, S.~Pan, F.~Chen, G.~Long, C.~Zhang, and P.~S. Yu, ``{A Comprehensive Survey on Graph Neural Networks},'' \emph{IEEE TNNLS}, vol.~32, no.~1, pp. 4--24, 2021.

\bibitem[Zhao et~al.(2020)Zhao, Song, Zhang, Liu, Wang, Lin, Deng, and Li]{T-GCN}
L.~Zhao, Y.~Song, C.~Zhang, Y.~Liu, P.~Wang, T.~Lin, M.~Deng, and H.~Li, ``{T-GCN: A Temporal Graph Convolutional Network for Traffic Prediction},'' \emph{IEEE T-ITS}, vol.~21, no.~9, pp. 3848--3858, 2020.

\bibitem[Rossi et~al.(2020)Rossi, Chamberlain, Frasca, Eynard, Monti, and Bronstein]{tgn_rossi2020}
E.~Rossi, B.~Chamberlain, F.~Frasca, D.~Eynard, F.~Monti, and M.~Bronstein, ``{Temporal Graph Networks for Deep Learning on Dynamic Graphs},'' in \emph{ICML 2020 Workshop on Graph Representation Learning}, 2020.

\bibitem[Trivedi et~al.(2019)Trivedi, Farajtabar, Biswal, and Zha]{dyrep}
R.~Trivedi, M.~Farajtabar, P.~Biswal, and H.~Zha, ``{DyRep: Learning Representations over Dynamic Graphs},'' in \emph{ICLR}, 2019.

\bibitem[Xu et~al.(2020)Xu, Ruan, Korpeoglu, Kumar, and Achan]{TGAT}
D.~Xu, C.~Ruan, E.~Korpeoglu, S.~Kumar, and K.~Achan, ``{Inductive representation learning on temporal graphs},'' in \emph{ICLR}, 2020.

\bibitem[Kazemi et~al.(2020)Kazemi, Goel, Jain, Kobyzev, Sethi, Forsyth, and Poupart]{dynamicgraph_survey}
S.~M. Kazemi, R.~Goel, K.~Jain, I.~Kobyzev, A.~Sethi, P.~Forsyth, and P.~Poupart, ``{Representation Learning for Dynamic Graphs: A Survey},'' \emph{J. Mach. Learn. Res.}, vol.~21, no.~1, 2020.

\bibitem[Jiang and Luo(2022)]{traffic_forecasting_survey}
W.~Jiang and J.~Luo, ``{Graph neural network for traffic forecasting: A survey},'' \emph{Expert Systems with Applications}, vol. 207, p. 117921, 2022.

\bibitem[Bondy(1976)]{graph_theory}
J.~A. Bondy, \emph{{Graph Theory With Applications}}.\hskip 1em plus 0.5em minus 0.4em\relax Elsevier Science Ltd., 1976.

\bibitem[Scarselli et~al.(2009)Scarselli, Gori, Tsoi, Hagenbuchner, and Monfardini]{GNN}
F.~Scarselli, M.~Gori, A.~C. Tsoi, M.~Hagenbuchner, and G.~Monfardini, ``{The Graph Neural Network Model},'' \emph{IEEE Transactions on Neural Networks}, vol.~20, no.~1, pp. 61--80, 2009.

\bibitem[Micheli(2009)]{NN4G}
A.~Micheli, ``{Neural Network for Graphs: A Contextual Constructive Approach},'' \emph{IEEE Transactions on Neural Networks}, vol.~20, no.~3, pp. 498--511, 2009.

\bibitem[Defferrard et~al.(2016)Defferrard, Bresson, and Vandergheynst]{chebnet}
M.~Defferrard, X.~Bresson, and P.~Vandergheynst, ``{Convolutional Neural Networks on Graphs with Fast Localized Spectral Filtering},'' in \emph{Proc. of the 29th NeurIPS}.\hskip 1em plus 0.5em minus 0.4em\relax Curran Associates Inc., 2016, p. 3844–3852.

\bibitem[Kipf and Welling(2017)]{GCN}
T.~N. Kipf and M.~Welling, ``{Semi-Supervised Classification with Graph Convolutional Networks},'' in \emph{ICLR}, 2017.

\bibitem[Veli{\v{c}}kovi{\'{c}} et~al.(2018)Veli{\v{c}}kovi{\'{c}}, Cucurull, Casanova, Romero, Li{\`{o}}, and Bengio]{GAT}
P.~Veli{\v{c}}kovi{\'{c}}, G.~Cucurull, A.~Casanova, A.~Romero, P.~Li{\`{o}}, and Y.~Bengio, ``{Graph Attention Networks},'' \emph{ICLR}, 2018.

\bibitem[Hamilton et~al.(2017{\natexlab{a}})Hamilton, Ying, and Leskovec]{SAGE}
W.~L. Hamilton, R.~Ying, and J.~Leskovec, ``{Inductive Representation Learning on Large Graphs},'' in \emph{NeurIPS}, 2017.

\bibitem[Chiang et~al.(2019)Chiang, Liu, Si, Li, Bengio, and Hsieh]{clusterGCN}
W.-L. Chiang, X.~Liu, S.~Si, Y.~Li, S.~Bengio, and C.-J. Hsieh, ``{Cluster-GCN: An Efficient Algorithm for Training Deep and Large Graph Convolutional Networks},'' in \emph{Proc. of the 25th ACM SIGKDD KDD}.\hskip 1em plus 0.5em minus 0.4em\relax Association for Computing Machinery, 2019, p. 257–266.

\bibitem[Xu et~al.(2019)Xu, Hu, Leskovec, and Jegelka]{GIN}
K.~Xu, W.~Hu, J.~Leskovec, and S.~Jegelka, ``{How Powerful are Graph Neural Networks?}'' in \emph{ICLR}, 2019.

\bibitem[Weisfeiler and Lehman(1968)]{WL}
B.~Weisfeiler and A.~Lehman, ``{A Reduction of a Graph to a Canonical Form and an Algebra Arising during This Reduction},'' \emph{Nauchno-Technicheskaya Informatsia}, vol.~2, no.~9, 1968.

\bibitem[Gravina et~al.(2023)Gravina, Bacciu, and Gallicchio]{gravina2023adgn}
A.~Gravina, D.~Bacciu, and C.~Gallicchio, ``Anti-{S}ymmetric {DGN}: a stable architecture for {D}eep {G}raph {N}etworks,'' in \emph{ICLR}, 2023.

\bibitem[Wang et~al.(2021{\natexlab{a}})Wang, Wang, Yang, and Lin]{dgc}
Y.~Wang, Y.~Wang, J.~Yang, and Z.~Lin, ``{Dissecting the Diffusion Process in Linear Graph Convolutional Networks},'' in \emph{NeurIPS}, 2021.

\bibitem[Wu et~al.(2019)Wu, Souza, Zhang, Fifty, Yu, and Weinberger]{sgc}
F.~Wu, A.~Souza, T.~Zhang, C.~Fifty, T.~Yu, and K.~Weinberger, ``{Simplifying Graph Convolutional Networks},'' in \emph{Proc. of the 36th ICML}, vol.~97.\hskip 1em plus 0.5em minus 0.4em\relax PMLR, 2019, pp. 6861--6871.

\bibitem[Eliasof et~al.(2021)Eliasof, Haber, and Treister]{pde-gcn}
M.~Eliasof, E.~Haber, and E.~Treister, ``{PDE-GCN: Novel Architectures for Graph Neural Networks Motivated by Partial Differential Equations},'' in \emph{NeurIPS}, 2021.

\bibitem[Rusch et~al.(2022)Rusch, Chamberlain, Rowbottom, Mishra, and Bronstein]{graphcon}
T.~K. Rusch, B.~P. Chamberlain, J.~Rowbottom, S.~Mishra, and M.~M. Bronstein, ``{Graph-Coupled Oscillator Networks},'' \emph{arXiv preprint arXiv:2202.02296}, 2022.

\bibitem[Perozzi et~al.(2014)Perozzi, Al-Rfou, and Skiena]{DeepWalk}
B.~Perozzi, R.~Al-Rfou, and S.~Skiena, ``{DeepWalk: Online Learning of Social Representations},'' in \emph{Proc. of the 20th ACM SIGKDD KDD}.\hskip 1em plus 0.5em minus 0.4em\relax ACM, 2014, pp. 701--710.

\bibitem[Mikolov et~al.(2013)Mikolov, Chen, Corrado, and Dean]{skipgram}
T.~Mikolov, K.~Chen, G.~Corrado, and J.~Dean, ``{Efficient Estimation of Word Representations in Vector Space},'' in \emph{ICLR}, 2013.

\bibitem[Grover and Leskovec(2016)]{node2vec}
A.~Grover and J.~Leskovec, ``{node2vec: Scalable Feature Learning for Networks},'' in \emph{Proc. of the 22nd ACM SIGKDD KDD}, 2016.

\bibitem[Gao et~al.(2010)Gao, Xiao, Tao, and Li]{Gao2010}
X.~Gao, B.~Xiao, D.~Tao, and X.~Li, ``A survey of graph edit distance,'' \emph{Pattern Analysis and Applications}, vol.~13, no.~1, pp. 113--129, 2010.

\bibitem[Paassen et~al.(2021)Paassen, Grattarola, Zambon, Alippi, and Hammer]{paassen2021graph}
B.~Paassen, D.~Grattarola, D.~Zambon, C.~Alippi, and B.~E. Hammer, ``Graph edit networks,'' in \emph{Proc. of ICLR 2021}, 2021.

\bibitem[Rumelhart et~al.(1986)Rumelhart, Hinton, and Williams]{RNN}
D.~E. Rumelhart, G.~E. Hinton, and R.~J. Williams, ``{Learning representations by back-propagating errors},'' \emph{Nature}, vol. 323, no. 6088, pp. 533--536, 1986.

\bibitem[Seo et~al.(2018)Seo, Defferrard, Vandergheynst, and Bresson]{GCRN}
Y.~Seo, M.~Defferrard, P.~Vandergheynst, and X.~Bresson, ``{Structured Sequence Modeling with Graph Convolutional Recurrent Networks},'' in \emph{NeurIPS}, 2018.

\bibitem[Gers et~al.(2002)Gers, Schraudolph, and Schmidhuber]{peephole-LSTM1}
F.~Gers, N.~Schraudolph, and J.~Schmidhuber, ``{Learning Precise Timing with LSTM Recurrent Networks},'' \emph{Journal of Machine Learning Research}, vol.~3, pp. 115--143, 2002.

\bibitem[Graves(2013)]{peephole-LSTM2}
A.~Graves, ``{Generating sequences with recurrent neural networks},'' \emph{arXiv preprint arXiv:1308.0850}, 2013.

\bibitem[Li et~al.(2018)Li, Yu, Shahabi, and Liu]{DCRNN}
Y.~Li, R.~Yu, C.~Shahabi, and Y.~Liu, ``{Diffusion Convolutional Recurrent Neural Network: Data-Driven Traffic Forecasting},'' in \emph{ICLR}, 2018.

\bibitem[Cho et~al.(2014)Cho, van Merrienboer, Gulcehre, Bahdanau, Bougares, Schwenk, and Bengio]{GRU}
K.~Cho, B.~van Merrienboer, C.~Gulcehre, D.~Bahdanau, F.~Bougares, H.~Schwenk, and Y.~Bengio, ``{Learning Phrase Representations using RNN Encoder-Decoder for Statistical Machine Translation},'' \emph{arXiv preprint arXiv:1406.1078}, 2014.

\bibitem[Bai et~al.(2021)Bai, Zhu, Song, Zhao, Hou, Du, and Li]{a3tgcn}
J.~Bai, J.~Zhu, Y.~Song, L.~Zhao, Z.~Hou, R.~Du, and H.~Li, ``{A3T-GCN}: {A}ttention {T}emporal {G}raph {C}onvolutional {N}etwork for {T}raffic {F}orecasting,'' \emph{ISPRS International Journal of Geo-Information}, vol.~10, no.~7, 2021.

\bibitem[Yu et~al.(2018)Yu, Yin, and Zhu]{STGCN}
B.~Yu, H.~Yin, and Z.~Zhu, ``{Spatio-temporal Graph Convolutional Networks: A Deep Learning Framework for Traffic Forecasting},'' in \emph{Proc. of the 27th IJCAI}, 2018.

\bibitem[Dauphin et~al.(2017)Dauphin, Fan, Auli, and Grangier]{GLU}
Y.~N. Dauphin, A.~Fan, M.~Auli, and D.~Grangier, ``{Language Modeling with Gated Convolutional Networks},'' in \emph{Proc. of the 34th ICML}, vol.~70.\hskip 1em plus 0.5em minus 0.4em\relax JMLR, 2017, p. 933–941.

\bibitem[Guo et~al.(2019)Guo, Lin, Feng, Song, and Wan]{astgcn}
S.~Guo, Y.~Lin, N.~Feng, C.~Song, and H.~Wan, ``Attention based spatial-temporal graph convolutional networks for traffic flow forecasting,'' \emph{Proc. of the AAAI Conference on Artificial Intelligence}, vol.~33, no.~01, pp. 922--929, 2019.

\bibitem[Chen et~al.(2018)Chen, Xu, Wu, and Zheng]{GC-LSTM}
J.~Chen, X.~Xu, Y.~Wu, and H.~Zheng, ``{GC-LSTM}: Graph convolution embedded lstm for dynamic link prediction,'' \emph{arXiv preprint arXiv:1812.04206}, 2018.

\bibitem[Li et~al.(2019)Li, Han, Cheng, Su, Wang, Zhang, and Pan]{LRGCN}
J.~Li, Z.~Han, H.~Cheng, J.~Su, P.~Wang, J.~Zhang, and L.~Pan, ``{Predicting Path Failure In Time-Evolving Graphs},'' in \emph{Proc. of the 25th ACM SIGKDD KDD}.\hskip 1em plus 0.5em minus 0.4em\relax Association for Computing Machinery, 2019, p. 1279–1289.

\bibitem[Schlichtkrull et~al.(2018)Schlichtkrull, Kipf, Bloem, van den Berg, Titov, and Welling]{RGCN}
M.~Schlichtkrull, T.~N. Kipf, P.~Bloem, R.~van den Berg, I.~Titov, and M.~Welling, ``{Modeling Relational Data with Graph Convolutional Networks},'' in \emph{The Semantic Web}.\hskip 1em plus 0.5em minus 0.4em\relax Springer International Publishing, 2018, pp. 593--607.

\bibitem[Micheli and Tortorella(2022)]{dyngesn}
A.~Micheli and D.~Tortorella, ``Discrete-time dynamic graph echo state networks,'' \emph{Neurocomputing}, vol. 496, pp. 85--95, 2022.

\bibitem[Gallicchio and Micheli(2010)]{gesn}
C.~Gallicchio and A.~Micheli, ``Graph echo state networks,'' in \emph{IJCNN}.\hskip 1em plus 0.5em minus 0.4em\relax IEEE, 2010, pp. 1--8.

\bibitem[Panagopoulos et~al.(2021)Panagopoulos, Nikolentzos, and Vazirgiannis]{mpnn_lsltm}
G.~Panagopoulos, G.~Nikolentzos, and M.~Vazirgiannis, ``{Transfer Graph Neural Networks for Pandemic Forecasting},'' in \emph{Proc. of the 35th AAAI Conference on Artificial Intelligence}, 2021.

\bibitem[You et~al.(2022)You, Du, and Leskovec]{roland}
J.~You, T.~Du, and J.~Leskovec, ``{ROLAND: graph learning framework for dynamic graphs},'' in \emph{Proc. of the 28th ACM SIGKDD KDD}, 2022, pp. 2358--2366.

\bibitem[Deng et~al.(2019)Deng, Rangwala, and Ning]{10.1145/3292500.3330919}
S.~Deng, H.~Rangwala, and Y.~Ning, ``Learning dynamic context graphs for predicting social events,'' in \emph{Proc. of the 25th ACM SIGKDD KDD}.\hskip 1em plus 0.5em minus 0.4em\relax Association for Computing Machinery, 2019, p. 1007–1016.

\bibitem[Cini et~al.(2023)Cini, Marisca, Bianchi, and Alippi]{cini2023scalable}
A.~Cini, I.~Marisca, F.~Bianchi, and C.~Alippi, ``{Scalable Spatiotemporal Graph Neural Networks},'' in \emph{Proc. of the AAAI Conference on Artificial Intelligence}, 2023.

\bibitem[Jaeger(2010)]{esn1}
H.~Jaeger, ``The “echo state” approach to analysing and training recurrent neural networks--with an erratum note,'' \emph{German National Research Center for Information Technology GMD Technical Report}, vol. 148, no.~34, 2010.

\bibitem[Jaeger and Haas(2004)]{esn2}
H.~Jaeger and H.~Haas, ``{Harnessing Nonlinearity: Predicting Chaotic Systems and Saving Energy in Wireless Communication},'' \emph{Science}, vol. 304, no. 5667, pp. 78--80, 2004.

\bibitem[Gallicchio and Scardapane(2020)]{randomizedNN}
C.~Gallicchio and S.~Scardapane, ``Deep randomized neural networks,'' in \emph{Recent Trends in Learning From Data: Tutorials from the INNS Big Data and Deep Learning Conference}.\hskip 1em plus 0.5em minus 0.4em\relax Springer, 2020, pp. 43--68.

\bibitem[Pareja et~al.(2020)Pareja, Domeniconi, Chen, Ma, Suzumura, Kanezashi, Kaler, Schardl, and Leiserson]{egcn}
A.~Pareja, G.~Domeniconi, J.~Chen, T.~Ma, T.~Suzumura, H.~Kanezashi, T.~Kaler, T.~B. Schardl, and C.~E. Leiserson, ``{EvolveGCN: Evolving Graph Convolutional Networks for Dynamic Graphs},'' in \emph{Proc. of the 34th AAAI Conference on Artificial Intelligence}, 2020.

\bibitem[Taheri and Berger-Wolf(2019)]{dygrae}
A.~Taheri and T.~Berger-Wolf, ``{Predictive Temporal Embedding of Dynamic Graphs},'' in \emph{Proc. of the 2019 IEEE/ACM ASONAM}.\hskip 1em plus 0.5em minus 0.4em\relax Association for Computing Machinery, 2019, p. 57–64.

\bibitem[Li et~al.(2016)Li, Zemel, Brockschmidt, and Tarlow]{gatedGNN}
Y.~Li, R.~Zemel, M.~Brockschmidt, and D.~Tarlow, ``{Gated Graph Sequence Neural Networks},'' in \emph{ICLR}, 2016.

\bibitem[Goyal et~al.(2020)Goyal, Chhetri, and Canedo]{dyngraph2vec}
P.~Goyal, S.~R. Chhetri, and A.~Canedo, ``dyngraph2vec: Capturing network dynamics using dynamic graph representation learning,'' \emph{Knowledge-Based Systems}, vol. 187, p. 104816, 2020.

\bibitem[Goyal et~al.(2018)Goyal, Kamra, He, and Liu]{dyngem}
P.~Goyal, N.~Kamra, X.~He, and Y.~Liu, ``{DynGEM: Deep Embedding Method for Dynamic Graphs},'' \emph{arXiv preprint arXiv:1805.11273}, 2018.

\bibitem[Bastas et~al.(2019)Bastas, Semertzidis, Axenopoulos, and Daras]{evolve2vec}
N.~Bastas, T.~Semertzidis, A.~Axenopoulos, and P.~Daras, ``{evolve2vec: Learning Network Representations Using Temporal Unfolding},'' in \emph{MultiMedia Modeling}.\hskip 1em plus 0.5em minus 0.4em\relax Springer International Publishing, 2019, pp. 447--458.

\bibitem[Kumar et~al.(2019)Kumar, Zhang, and Leskovec]{jodie}
S.~Kumar, X.~Zhang, and J.~Leskovec, ``{Predicting Dynamic Embedding Trajectory in Temporal Interaction Networks},'' in \emph{Proc. of the 25th ACM SIGKDD KDD}, 2019.

\bibitem[Ma et~al.(2020)Ma, Guo, Ren, Tang, and Yin]{streamgnn}
Y.~Ma, Z.~Guo, Z.~Ren, J.~Tang, and D.~Yin, ``{Streaming Graph Neural Networks},'' in \emph{Proc. of the 43rd International ACM SIGIR}.\hskip 1em plus 0.5em minus 0.4em\relax Association for Computing Machinery, 2020, p. 719–728.

\bibitem[Nguyen et~al.(2018)Nguyen, Lee, Rossi, Ahmed, Koh, and Kim]{temporal_node2vec}
G.~Nguyen, J.~B. Lee, R.~A. Rossi, N.~Ahmed, E.~Koh, and S.~Kim, ``{Continuous-Time Dynamic Network Embeddings},'' \emph{Companion Proc. of the Web Conference}, 2018.

\bibitem[Wang et~al.(2021{\natexlab{b}})Wang, Chang, Liu, Leskovec, and Li]{CAW}
Y.~Wang, Y.-Y. Chang, Y.~Liu, J.~Leskovec, and P.~Li, ``{Inductive Representation Learning in Temporal Networks via Causal Anonymous Walks},'' in \emph{ICLR}, 2021.

\bibitem[Jin et~al.(2022)Jin, Li, and Pan]{neurtws}
M.~Jin, Y.-F. Li, and S.~Pan, ``{Neural Temporal Walks: Motif-Aware Representation Learning on Continuous-Time Dynamic Graphs},'' in \emph{NeurIPS}, 2022.

\bibitem[Souza et~al.(2022)Souza, Mesquita, Kaski, and Garg]{pint}
A.~H. Souza, D.~Mesquita, S.~Kaski, and V.~K. Garg, ``Provably expressive temporal graph networks,'' in \emph{NeurIPS}, 2022.

\bibitem[Errica et~al.(2023)Errica, Bacciu, and Micheli]{pydgn}
F.~Errica, D.~Bacciu, and A.~Micheli, ``{PyDGN: a Python Library for Flexible and Reproducible Research on Deep Learning for Graphs},'' \emph{Journal of Open Source Software}, vol.~8, no.~90, 2023.

\bibitem[Fey and Lenssen(2019)]{Fey/Lenssen/2019}
M.~Fey and J.~E. Lenssen, ``Fast graph representation learning with {PyTorch Geometric},'' in \emph{ICLR Workshop on Representation Learning on Graphs and Manifolds}, 2019.

\bibitem[Leskovec and Krevl(2014)]{snapnets}
J.~Leskovec and A.~Krevl, ``{SNAP Datasets}: {Stanford} large network dataset collection,'' \url{http://snap.stanford.edu/data}, 2014.

\bibitem[Cini and Marisca()]{tsl}
A.~Cini and I.~Marisca, ``Torch spatiotemporal, 3 2022,'' \emph{\url{https://github.com/TorchSpatiotemporal/tsl}}, vol.~10.

\bibitem[Huang et~al.(2023)Huang, Poursafaei, Danovitch, Fey, Hu, Rossi, Leskovec, Bronstein, Rabusseau, and Rabbany]{TGB}
S.~Huang, F.~Poursafaei, J.~Danovitch, M.~Fey, W.~Hu, E.~Rossi, J.~Leskovec, M.~M. Bronstein, G.~Rabusseau, and R.~Rabbany, ``{Temporal Graph Benchmark for Machine Learning on Temporal Graphs},'' in \emph{NeurIPS Datasets and Benchmarks Track}, 2023.

\bibitem[Rossi and Ahmed(2015)]{nr}
\BIBentryALTinterwordspacing
R.~A. Rossi and N.~K. Ahmed, ``The network data repository with interactive graph analytics and visualization,'' in \emph{AAAI}, 2015. [Online]. Available: \url{https://networkrepository.com}
\BIBentrySTDinterwordspacing

\bibitem[Rozemberczki et~al.(2021)Rozemberczki, Scherer, He, Panagopoulos, Riedel, Astefanoaei, Kiss, Beres, Lopez, Collignon, and Sarkar]{rozemberczki2021pytorch}
B.~Rozemberczki, P.~Scherer, Y.~He, G.~Panagopoulos, A.~Riedel, M.~Astefanoaei, O.~Kiss, F.~Beres, G.~Lopez, N.~Collignon, and R.~Sarkar, ``{PyTorch Geometric Temporal: Spatiotemporal Signal Processing with Neural Machine Learning Models},'' in \emph{Proc. of the 30th ACM CIKM}, 2021, p. 4564–4573.

\bibitem[B{\'e}res et~al.(2018)B{\'e}res, P{\'a}lovics, Ol{\'a}h, and Bencz{\'u}r]{twitter_tennis}
F.~B{\'e}res, R.~P{\'a}lovics, A.~Ol{\'a}h, and A.~A. Bencz{\'u}r, ``Temporal walk based centrality metric for graph streams,'' \emph{Applied Network Science}, vol.~3, no.~1, p.~32, 2018.

\bibitem[Weber et~al.(2019)Weber, Domeniconi, Chen, Weidele, Bellei, Robinson, and Leiserson]{elliptic}
M.~Weber, G.~Domeniconi, J.~Chen, D.~Weidele, C.~Bellei, T.~Robinson, and C.~Leiserson, ``{Anti-Money Laundering in Bitcoin: Experimenting with Graph Convolutional Networks for Financial Forensics},'' \emph{KDD ’19 Workshop on Anomaly Detection in Finance}, 2019.

\bibitem[Leskovec et~al.(2005)Leskovec, Kleinberg, and Faloutsos]{as733}
J.~Leskovec, J.~Kleinberg, and C.~Faloutsos, ``{Graphs over Time: Densification Laws, Shrinking Diameters and Possible Explanations},'' in \emph{Proc. of the 11th ACM SIGKDD KDD}.\hskip 1em plus 0.5em minus 0.4em\relax Association for Computing Machinery, 2005, p. 177–187.

\bibitem[Kumar et~al.(2016)Kumar, Spezzano, Subrahmanian, and Faloutsos]{bc-otc}
S.~Kumar, F.~Spezzano, V.~Subrahmanian, and C.~Faloutsos, ``{Edge weight prediction in weighted signed networks},'' in \emph{IEEE ICDM}.\hskip 1em plus 0.5em minus 0.4em\relax IEEE, 2016, pp. 221--230.

\bibitem[Kumar et~al.(2018)Kumar, Hooi, Makhija, Kumar, Faloutsos, and Subrahmanian]{bc-otc2}
S.~Kumar, B.~Hooi, D.~Makhija, M.~Kumar, C.~Faloutsos, and V.~Subrahmanian, ``{Rev2: Fraudulent user prediction in rating platforms},'' in \emph{Proc. of the 11th ACM WSDM}.\hskip 1em plus 0.5em minus 0.4em\relax ACM, 2018, pp. 333--341.

\bibitem[Poursafaei et~al.(2022)Poursafaei, Huang, Pelrine, , and Rabbany]{edgebank}
F.~Poursafaei, S.~Huang, K.~Pelrine, , and R.~Rabbany, ``Towards better evaluation for dynamic link prediction,'' in \emph{NeurIPS Datasets and Benchmarks}, 2022.

\bibitem[Pei et~al.(2020)Pei, Wei, Chang, Lei, and Yang]{geom-gcn}
H.~Pei, B.~Wei, K.~C.-C. Chang, Y.~Lei, and B.~Yang, ``{Geom-GCN: Geometric Graph Convolutional Networks},'' in \emph{ICLR}, 2020.

\bibitem[Yan et~al.(2022)Yan, Hashemi, Swersky, Yang, and Koutra]{heterophily_results2}
Y.~Yan, M.~Hashemi, K.~Swersky, Y.~Yang, and D.~Koutra, ``{Two Sides of the Same Coin: Heterophily and Oversmoothing in Graph Convolutional Neural Networks},'' in \emph{IEEE ICDM}, 2022, pp. 1287--1292.

\bibitem[Cavallo et~al.(2023)Cavallo, Grohnfeldt, Russo, Lovisotto, and Vassio]{cavallo2023gcnh}
A.~Cavallo, C.~Grohnfeldt, M.~Russo, G.~Lovisotto, and L.~Vassio, ``{GCNH: A Simple Method For Representation Learning On Heterophilous Graphs},'' \emph{arXiv preprint arXiv:2304.10896}, 2023.

\bibitem[Ji et~al.(2022)Ji, Pan, Cambria, Marttinen, and Yu]{heterogeneous}
S.~Ji, S.~Pan, E.~Cambria, P.~Marttinen, and P.~S. Yu, ``{A Survey on Knowledge Graphs: Representation, Acquisition, and Applications},'' \emph{IEEE TNNLS}, vol.~33, no.~2, pp. 494--514, 2022.

\bibitem[Li et~al.(2022)Li, Liu, Zhang, Liu, and Xiong]{heterogeneous2}
Z.~Li, H.~Liu, Z.~Zhang, T.~Liu, and N.~N. Xiong, ``{Learning Knowledge Graph Embedding With Heterogeneous Relation Attention Networks},'' \emph{IEEE TNNLS}, vol.~33, no.~8, pp. 3961--3973, 2022.

\bibitem[Maddalena et~al.(2022)Maddalena, Giordano, Manzo, and Guarracino]{Maddalena2022}
L.~Maddalena, M.~Giordano, M.~Manzo, and M.~R. Guarracino, \emph{Whole-Graph Embedding and Adversarial Attacks for Life Sciences}.\hskip 1em plus 0.5em minus 0.4em\relax Cham: Springer International Publishing, 2022, pp. 1--21.

\bibitem[Deng et~al.(2022)Deng, Lian, Huang, and Chen]{adversarial_graph}
L.~Deng, D.~Lian, Z.~Huang, and E.~Chen, ``{Graph Convolutional Adversarial Networks for Spatiotemporal Anomaly Detection},'' \emph{IEEE TNNLS}, vol.~33, no.~6, pp. 2416--2428, 2022.

\bibitem[Cai and Wang(2020)]{over-smoothing}
C.~Cai and Y.~Wang, ``A note on over-smoothing for graph neural networks,'' \emph{arXiv preprint arXiv:2006.13318}, 2020.

\bibitem[Alon and Yahav(2021)]{bottleneck}
U.~Alon and E.~Yahav, ``{On the Bottleneck of Graph Neural Networks and its Practical Implications},'' in \emph{ICLR}, 2021.

\bibitem[Giovanni et~al.(2023)Giovanni, Giusti, Barbero, Luise, Lio', and Bronstein]{digiovanni2023oversquashing}
F.~D. Giovanni, L.~Giusti, F.~Barbero, G.~Luise, P.~Lio', and M.~Bronstein, ``{On Over-Squashing in Message Passing Neural Networks: The Impact of Width, Depth, and Topology},'' \emph{arXiv preprint arXiv:2302.02941}, 2023.

\bibitem[Li and Leskovec(2022)]{GNNBook-ch5-li}
P.~Li and J.~Leskovec, ``The expressive power of graph neural networks,'' in \emph{Graph Neural Networks: Foundations, Frontiers, and Applications}, L.~Wu, P.~Cui, J.~Pei, and L.~Zhao, Eds.\hskip 1em plus 0.5em minus 0.4em\relax Springer Singapore, 2022, pp. 63--98.

\bibitem[Gan et~al.(2022)Gan, Hu, Mo, Kang, Peng, Zhu, and Zhu]{multigraph}
J.~Gan, R.~Hu, Y.~Mo, Z.~Kang, L.~Peng, Y.~Zhu, and X.~Zhu, ``{Multigraph Fusion for Dynamic Graph Convolutional Network},'' \emph{IEEE TNNLS}, pp. 1--12, 2022.

\bibitem[Hamilton et~al.(2017{\natexlab{b}})Hamilton, Ying, and Leskovec]{hamilton}
\BIBentryALTinterwordspacing
W.~L. Hamilton, R.~Ying, and J.~Leskovec, ``Representation learning on graphs: Methods and applications,'' \emph{{IEEE} Data Eng. Bull.}, vol.~40, no.~3, pp. 52--74, 2017. [Online]. Available: \url{http://sites.computer.org/debull/A17sept/p52.pdf}
\BIBentrySTDinterwordspacing

\end{thebibliography}

\end{document}